\documentclass[onecolumn,journal]{IEEEtran}
\usepackage{amsmath, amsfonts}
\usepackage{algorithmic}
\usepackage{algorithm}
\usepackage{cite}
\usepackage{array}
\usepackage{enumitem}
\usepackage[caption=false,font=normalsize,labelfont=sf,textfont=sf]{subfig}
\usepackage{textcomp}
\usepackage{url}
\usepackage{graphicx}
\usepackage{xcolor}
\usepackage{tikz}
\usetikzlibrary{trees,positioning,shapes,shadows,arrows.meta}
\usepackage{pgfplots}
\usepackage{adjustbox}
\usepackage{makecell}
\usepackage{longtable}
\usepackage{multirow}
\usepackage{tabularx}
\usepackage{forest}
\useforestlibrary{edges}
\pgfplotsset{compat=1.18}
\usetikzlibrary{positioning, decorations.pathreplacing}
\usepackage{balance}
\usepackage{afterpage}
\usepackage{tikz}
\usetikzlibrary{arrows.meta,positioning}
\usepackage{xcolor}

\definecolor{year2020}{RGB}{0,102,204}   
\definecolor{year2021}{RGB}{204,0,0}     
\definecolor{year2022}{RGB}{0,153,0}     
\definecolor{year2023}{RGB}{153,0,153}   
\definecolor{year2024}{RGB}{255,128,0} 
\definecolor{year2025}{RGB}{0,128,128} 

\tikzset{
    arrow/.style={-Stealth, very thick},
    box2020/.style={draw=year2020, text=year2020, rounded corners,
                    minimum width=2.8cm, minimum height=0.20cm},
    box2021/.style={draw=year2021, text=year2021, rounded corners,
                    minimum width=2.8cm, minimum height=0.20cm},
    box2022/.style={draw=year2022, text=year2022, rounded corners,
                    minimum width=2.8cm, minimum height=0.20cm},
    box2023/.style={draw=year2023, text=year2023, rounded corners,
                    minimum width=2.8cm, minimum height=0.20cm},
    box2024/.style={draw=year2024, text=year2024, rounded corners,
                    minimum width=2.8cm, minimum height=0.20cm},
    box2025/.style={draw=year2025, text=year2025, rounded corners,
                    minimum width=2.8cm, minimum height=0.20cm},
}

\begin{document}

\title{Intelligent Offloading in Vehicular Edge Computing: A Comprehensive Review of Deep Reinforcement Learning Approaches and Architectures}

\author{Ashab Uddin,~\IEEEmembership{Graduate Student Member,~IEEE}, Ning Zhang,~\IEEEmembership{Senior Member,~IEEE}, and Ahmed Hamdi Sakr,~\IEEEmembership{Senior Member,~IEEE}
\thanks{A. Uddin, N. Zhang, and A. H. Sakr are with the Department of Electrical and Computer Engineering, University of Windsor, Windsor, ON N9B 3P4, Canada. Email: \{uddin81, ahmed.sakr, ning.zhang\}@uwindsor.ca.}
}



\maketitle


\begin{abstract}
The increasing complexity of Intelligent Transportation Systems (ITS) has led to significant interest in computational task offloading to external infrastructures such as edge servers, vehicular nodes, and unmanned aerial vehicles (UAV) as well as supportive architectures such as Vehicular Edge Computing (VEC) networks. The dynamic and heterogeneous environments in VEC pose challenges for traditional offloading strategies, prompting the exploration of Reinforcement Learning (RL) and Deep Reinforcement Learning (DRL) as adaptive decision-making frameworks. This survey presents a comprehensive review of recent advances in DRL-based offloading for VEC. We classify and compare existing works based on learning paradigms (e.g., single-agent, multi-agent), system architectures (e.g., centralized, distributed, hierarchical), and optimization objectives (e.g., latency, energy, fairness). Furthermore, we analyze how Markov Decision Process (MDP) formulations are applied and highlight emerging trends in reward design, coordination mechanisms, and scalability. Finally, we identify open challenges and outline future research directions to guide the development of robust and intelligent offloading strategies for next-generation ITS.
\end{abstract}


\section{Introduction}
\subsection{Background and Motivation}
The automotive industry has undergone a transformative shift with the integration of embedded sensors, onboard processing units, and wireless communication technologies. These advancements have enabled the deployment of data-driven applications that enhance vehicle safety, efficiency, and comfort \cite{b201}. As a result, vehicular systems now generate substantial volumes of data and require increasingly complex computations, which are characterized by high delay sensitivity and strict real-time constraints. Applications such as autonomous driving, real-time traffic control, and advanced driver-assistance systems (ADAS) exemplify this growing demand for computational resources and low-latency processing. However, the dynamic and mobile nature of vehicular environments presents several operational challenges. Vehicles operate under varying channel conditions, fluctuating workloads, and constrained onboard computational and energy resources \cite{b2021}. These limitations hinder their ability to process tasks locally, necessitating external computational support.

Computation offloading addresses this challenge by enabling vehicles to delegate resource-intensive tasks to more powerful computing infrastructures, including centralized cloud computing (CC), Mobile Edge Computing (MEC), and fog nodes \cite{b202,b203,ba07,ba06,b204}. Among these, MEC has gained significant attention due to its ability to deploy computation and storage resources at the network edge, closer to the vehicles, using Roadside Units (RSUs), onboard modules, and infrastructure such as parking areas and gas stations \cite{b207}. This proximity reduces end-to-end latency and supports the stringent reliability requirements of 5G and emerging 6G networks \cite{b208,b2081}. In parallel, fog computing provides an intermediate coordination layer, enabling collaboration among edge nodes and facilitating scalable, localized resource management \cite{b209}.

Vehicular systems rely on Vehicle-to-Vehicle (V2V) and Vehicle-to-Infrastructure (V2I) communication paradigms to support connectivity between vehicles and edge infrastructures. These are implemented using radio access technologies such as Dedicated Short-Range Communications (DSRC), LTE, and 5G New Radio (NR), enabling robust and low-latency data exchange in heterogeneous vehicular networks \cite{b210}. Nonetheless, service continuity remains a key concern in mobile scenarios. As vehicles move across different regions, static associations with edge servers can result in performance degradation or service disruption. Service migration mechanisms address this issue by dynamically relocating services to edge nodes closest to the user, maintaining Quality of Service (QoS) while balancing migration costs and preserving session continuity \cite{b211}.

The scalability and heterogeneity of vehicular networks further emphasize the need for intelligent resource management \cite{b101}. Efficient task scheduling and load balancing are essential to meet latency, energy, and throughput constraints. Emerging approaches increasingly integrate machine learning (ML), and particularly RL, to intelligently orchestrate offloading decisions, manage resource allocation, and predict system dynamics across multi-layered infrastructures involving edge, fog, cloud, and UAV platforms. These techniques improve system responsiveness and prevent resource contention, paving the way for more autonomous and scalable VEC solutions. However, existing studies and surveys i.e., \cite{b2074,b174,b176} often examine learning-based offloading solutions and lack a unified perspective that integrates system architectures, MDP modeling, and coordination across dynamic vehicular environments. To address this gap, this survey focuses on DRL-based computation offloading in vehicular edge computing, systematically analyzing architectural paradigms, MDP formulations, learning frameworks for dynamic and heterogeneous settings, as well as multi-agent collaboration, synchronization, and generalization mechanisms.

\subsection{Related Studies}
In recent years, edge computing and computational task offloading have emerged as essential solutions to address challenges in modern vehicular, mobile, and IoT networks. Researchers have explored various architectures such as MEC, ad hoc systems, UAV networks, and cloudlets to improve performance, reduce latency, and optimize resource management. Surveys such as those  \cite{b2071} and \cite{b152} emphasized MEC's role in latency reduction, resource allocation, and privacy enhancement, with applications in smart homes, augmented reality, and healthcare. Comparative studies like   \cite{b2072}  highlighted the strengths of hybrid Cloud-Edge models, while others, like  \cite{b1072}, \cite{b151}, and  \cite{b211}, stressed the need for advancements in resource allocation, Edge Intelligence, and blockchain integration for security. Studies on vehicular networks, including the work of  \cite{b204}, examined AI and RL for dynamic task optimization and the potential of digital twin edge networks, and \cite{b202} studied the feasibility of server deployment. Furthermore, vehicular fog computing  and heterogeneous vehicular networks have been explored to leverage vehicular resources and improve connectivity while the integration of MEC with Software-Defined Networking (SDN) and Network Function Virtualization (NFV), has been discussed by \cite{b1071, b207}, further highlighted advances in network performance and scalability, particularly in 5G environments.
Furthermore, the integration of distributed learning techniques, such as federated reinforcement learning~\cite{b207a}, enhances delay performance, energy efficiency, and task completion through collaborative offloading among vehicles. Additionally, topology-adaptive offloading methods based on graph reinforcement learning~\cite{b207b} utilize V2V and V2I connectivity structures to optimize computational task offloading paths in dynamic network environments.


Traditional optimization techniques such as convex optimization \cite{b83}, evolutionary algorithms \cite{b85}, and game-theory \cite{b88} approaches face significant challenges in solving the problem of mixed integer nonlinearity, along with the increasing size and complexity of vehicular networks. These methods often struggle with scalability, adaptability to dynamic environments, and computational efficiency, especially in systems with large state-action spaces and non-convex dynamics. In contrast, DRL offers a promising solution by simultaneously learning state transitions and optimizing decision-making policies in real time. DRL's ability to handle both centralized and multi-agent frameworks enables it to adapt to diverse vehicular scenarios, making it well-suited for dynamic and uncertain environments. 

Considering the solution perspective through DRL, \cite{b2073} highlighted DRL's potential for MEC-enabled aerial access networks and emphasized challenges of wireless power transfer, physical-layer security, reconfigurable intelligent surfaces (RIS), and advanced multiple access mechanisms such as Non-Orthogonal Multiple Access (NOMA) and Rate-Splitting Multiple Access (RSMA). The work in \cite{b2074} noted MDP limitations in IoT, advocating robust frameworks for decentralized and multi-agent systems. Meanwhile, \cite{b203} identified scalability and coordination gaps in vehicular networks, suggesting hierarchical DRL models for efficient offloading. Focusing on One of the key challenges that arise in vehicular networks, such as high mobility and the non-stationary nature of the environment, in  \cite{b175} detailed how multi-agent reinforcement learning (MARL) frameworks can provide robust solutions, offering examples of MARL's application to V2V, V2I, and V2X communications.  Another key challenge i.e edge caching, has been examined  in \cite{b176} illustrating how DRL can improve caching strategies in mobile networks by making intelligent, data-driven decisions based on user behavior and network conditions.  Finally, \cite{b174} presented a comprehensive survey of MARL in future Internet technologies, focusing on challenges like dynamic network access, transmit power control, and computation offloading. A progress map is shown in Fig. \ref{fig:vec_timeline_color} to illustrate how DRL has evolved in VEC. The timeline highlights the major stages and key shifts in the field, showing how it progressed from simpler, heuristic methods to more advanced multi-agent and hierarchical approaches.

\begin{figure*}[t]
\centering
\begin{tikzpicture}[font=\tiny]

\draw[very thick] (-1,0) -- (16,0);

\coordinate (y2020) at (0,0);
\draw[arrow,year2020, < -] (y2020) -- ++(0,2.7);

\node[year2020, above=2.7cm of y2020] {\textbf{2020}};

\node[box2020, anchor=west, minimum width=2cm] (b201)
    at ([yshift=2.6cm, xshift=0.5cm]y2020)
    {\cite{b175},\cite{b176} Earliest  RL/MARL survey with for VEC.};

\node[box2020, below=0.05cm of b201.south west, anchor=north west] (b202)
    {\cite{b7} Earliest Priority-Aware DRL Offloading};

\node[box2020, below=0.05cm of b202.south west, anchor=north west]  (b203)
    {\cite{b78}, \cite{b3} Earliest DRL Offloading Game};
\node[box2020, below=0.05cm of b203.south west, anchor=north west]  (b204)
    {\cite{b78a} Joint radio--compute resource management};
\node[box2020, below=0.05cm of b204.south west, anchor=north west]  (b205)
    {\cite{b15} Collaboration among vehicles, RSUs, and MEC };

\draw[<-, thick] (y2020 |- b201.west) -- (b201.west);
\draw[<-, thick] (y2020 |- b202.west) -- (b202.west);  
\draw[<-, thick] (y2020 |- b203.west) -- (b203.west); 
\draw[<-, thick] (y2020 |- b204.west) -- (b204.west); 
\draw[<-, thick] (y2020 |- b205.west) -- (b205.west); 


\coordinate (y2021) at (2.5,0);
\draw[arrow,year2021, < -] (y2021) -- ++(0,-2.7);

\node[year2021, below=2.7cm of y2021] {\textbf{2021}};

\node[box2021, anchor=west, minimum width=2cm] (b301)
    at ([yshift=-2.6cm, xshift=0.5cm]y2021)
    {\cite{b4} Digital Twin Integration };

\node[box2021, above=0.05cm of b301.north west, anchor=south west] (b302)
    { \cite{b28} Asynchronous DRL with real trace-driven data };

\node[box2021, above=0.05cm of b302.north west, anchor=south west]  (b303)
    {\cite{b2} Blockchain-Based Secure Computation Offloading};
\node[box2021, above=0.05cm of b303.north west, anchor=south west]  (b304)
    {\cite{b5} Handover Service Integration};
\node[box2021, above=0.05cm of b304.north west, anchor=south west]  (b305)
    {\cite{b73} Game Combined Multiagent System Integration};

\draw[ < -, thick] (y2021 |- b301.west) -- (b301.west);
\draw[ < -, thick] (y2021 |- b302.west) -- (b302.west);  
\draw[ < -, thick] (y2021 |- b303.west) -- (b303.west);
\draw[ < -, thick] (y2021 |- b304.west) -- (b304.west);
\draw[ < -, thick] (y2021 |- b305.west) -- (b305.west);

\draw[<-, line width=0.75mm, shorten <=1.5mm, color=year2020]
    (y2021 |- y2020.west) -- (y2020.west);


\coordinate (y2022) at (5,0);
\draw[arrow,year2022, < -] (y2022) -- ++(0,2.7);

\node[year2022, above=2.7cm of y2022] {\textbf{2022}};

\node[box2022, anchor=west] (b401)
    at ([yshift=2.6cm, xshift=0.5cm]y2022)
    {\cite{b203},\cite{b204}, \cite{b1071} Landmark survey RL/DRL};

\node[box2022, below=0.05cm of b401.south west, anchor=north west] (b402)
    {\cite{b163}  Cooperative Data Sensing via UAV};

\node[box2022, below=0.05cm of b402.south west, anchor=north west]  (b403)
    {\cite{b31} Integration of Economic energy officient objectives};
\node[box2022, below=0.05cm of b403.south west, anchor=north west, minimum width=2cm]  (b404)
    {\cite{b34} Integration of NOMA};
\node[box2022, below=0.05cm of b404.south west, anchor=north west]  (b405)
    {\cite{b20} Integration of Vehicles Trust Value for security };

\draw[ < -, thick] (y2022 |- b401.west) -- (b401.west);
\draw[ < -, thick] (y2022 |- b402.west) -- (b402.west);  
\draw[ < -, thick] (y2022 |- b403.west) -- (b403.west); 
\draw[ < -, thick] (y2022 |- b404.west) -- (b404.west); 

\draw[ < -, thick] (y2022 |- b405.west) -- (b405.west);

\draw[<-, line width=1mm, shorten <=1.5mm,color=year2021]
    (y2022 |- y2021.west) -- (y2021.west);

\coordinate (y2023) at (7.5,0);
\draw[arrow,year2023,< -] (y2023) -- ++(0,-2.7);

\node[year2023, below=2.7cm of y2023] {\textbf{2023}};

\node[box2023, anchor=west] (b501)
    at ([yshift=-2.6cm, xshift=0.5cm]y2023)
    {\cite{b11} Introduction of improved  algorithm, Double DQN};

\node[box2023, above=0.05cm of b501.north west, anchor=south west] (b502)
    {\cite{b6} Introduced mobility prediction with DRL };

\node[box2023, above=0.05cm of b502.north west, anchor=south west]  (b503)
    {\cite{b55} DRL-Based Service Function Chain};
\node[box2023, above=0.05cm of b503.north west, anchor=south west]  (b504)
    {\cite{b21} User Scheduling for 8G network};
\node[box2023, above=0.05cm of b504.north west, anchor=south west]  (b505)
    {\cite{b35} Integration of Beamforming Optimization };

\draw[<-, thick] (y2023 |- b501.west) -- (b501.west);
\draw[<-, thick] (y2023 |- b502.west) -- (b502.west);  
\draw[<-, thick] (y2023 |- b503.west) -- (b503.west);
\draw[<-, thick] (y2023 |- b504.west) -- (b504.west);
\draw[<-, thick] (y2023 |- b505.west) -- (b505.west);

\draw[<-, line width=1.25mm, shorten <=1.5mm,color=year2022]
    (y2023 |- y2022.west) -- (y2022.west);
\coordinate (y2024) at (10,0);
\draw[arrow,year2024, < -] (y2024) -- ++(0,2.7);

\node[year2024, above=2.7cm of y2024] {\textbf{2024}};

\node[box2024, anchor=west] (b601)
    at ([yshift=2.6cm, xshift=0.5cm]y2024)
    {\cite{b76}  MA Cooperative UAV-
Assisted MEC};

\node[box2024, below=0.05cm of b601.south west, anchor=north west] (b602)
    {\cite{b52}, RIS-Enabled DRL Offload-
ing};

\node[box2024, below=0.05cm of b602.south west, anchor=north west]  (b603)
    {\cite{b62},\cite{b61},\cite{b72} Hiererchial Decision Making with DRL};
\node[box2024, below=0.05cm of b603.south west, anchor=north west]  (b604)
    {\cite{b207b} Integration of Federated Reinforcement Learning};
\node[box2024, below=0.05cm of b604.south west, anchor=north west]  (b605)
    {\cite{b82} Integration Prioritized user and task selection};

\draw[<-, thick] (y2024 |- b601.west) -- (b601.west);
\draw[<-, thick] (y2024 |- b602.west) -- (b602.west);  
\draw[<-, thick] (y2024 |- b603.west) -- (b603.west); 
\draw[<-, thick] (y2024 |- b604.west) -- (b604.west); 
\draw[<-, thick] (y2024 |- b605.west) -- (b605.west); 

\draw[<-, line width=1.5mm, shorten <=1.5mm,color=year2023]
    (y2024 |- y2023.west) -- (y2023.west);

\coordinate (y2025) at (12.5,0);
\draw[arrow,year2025,< -] (y2025) -- ++(0,-2.7);

\node[year2025, below=2.7cm of y2025] {\textbf{2025}};

\node[box2025, anchor=west] (b701)
    at ([yshift=-2.6cm, xshift=0.5cm]y2025)
    {\cite{b72a}  Transformer-Enhanced Distributed DRL};

\node[box2025, above=0.05cm of b701.north west, anchor=south west] (b702)
    {{\cite{b72b} Quantum Computing Integrated DRL}};

\node[box2025, above=0.05cm of b702.north west, anchor=south west]  (b703)
    {{\cite{b72c} Integration of Model with Model Free DRL}};

\draw[<-, thick] (y2025 |- b701.west) -- (b701.west);
\draw[<-, thick] (y2025 |- b702.west) -- (b702.west);  
\draw[<-, thick] (y2025 |- b703.west) -- (b703.west);

\draw[<-, line width=1.75mm, shorten <=1.5mm,color=year2024]
    (y2025 |- y2024.west) -- (y2024.west);

\coordinate (y2026) at (15,0);
\draw[->, line width=2mm, shorten >=1.5mm,color=year2025] 
    (y2025) -- (y2026);

\end{tikzpicture}

\caption{Timeline of DRL development in VEC}

\label{fig:vec_timeline_color}
\end{figure*}

\begin{table*}[t]
\centering
\caption{Limitations of Conventional Optimization and How DRL Addresses Them}
\label{tab:drl_motivation}
\renewcommand{\arraystretch}{1.2}
\setlength{\tabcolsep}{5pt}
\small
\begin{tabular}{|p{3.2cm}|p{5.8cm}|p{5.8cm}|}
\hline
\textbf{Aspect} & \textbf{Conventional Optimization Limitations} & \textbf{How DRL Overcomes the Limitation} \\
\hline
\hline
Long-term decision making & 
Cannot effectively capture long-term dependencies or delayed rewards due to myopic or short-horizon formulations. &
Optimizes long-term cumulative rewards using temporal-difference learning and neural networks over entire episodes. \\
\hline
High-dimensional state/action spaces &
Scales poorly with large state or action spaces, leading to prohibitive computational complexity. &
Employs deep neural networks for function approximation, enabling learning in high-dimensional and continuous spaces. \\
\hline
Mixed-integer decision variables &
Mixed discrete-continuous optimization problems are computationally expensive and often intractable. &
Supports hybrid action spaces through hierarchical policies, parameterized actions, or structured action representations. \\
\hline
Non-convex optimization &
Requires heuristics or approximations, with no guarantee of near-optimal solutions. &
Explores non-convex solution spaces using stochastic gradient-based optimization and experience replay. \\
\hline
Partial observability &
Typically assumes full observability and struggles with POMDP settings. &
Handles partial observability via belief states, history-based policies, multi-agent DRL architectures. \\
\hline
\end{tabular}
\end{table*}

To clarify the motivation and scope of this survey, Table~\ref{tab:drl_motivation} summarizes the limitations of conventional optimization approaches and highlights how DRL addresses these challenges, while Table~\ref{tab:rw_contribution_comparison} compares representative surveys and their scope relative to the unified perspective adopted in this work. Studies on DRL  highlights not only potential solutions for mixed integer nonlinear problems, but also demonstrates a greater ability to adapt and cope with varying conditions and large networks. However, a key challenge that remains under-explored in current DRL research is the examination of MDP formulations with learning methods that fully account for the complexities of dynamic environments while meeting system-wide goals. Effective coordination of MDP formulations with state-action transitions and reward structures tailored to specific DRL approaches is essential for optimizing performance. This includes designing reward functions that not only guide agents toward local objectives but also ensure overall system efficiency. In multi-agent systems, uncoordinated individual decisions often lead to suboptimal outcomes, such as resource contention and inefficiency. Furthermore, aggregating individual Partially Observable MDPs  (POMDPs), into a coherent global MDP is complex where agents cannot directly observe the true environment state, requiring an understanding of how these MDPs interact, synchronize, or merge to form a valid and accurate representation of the whole environment. Studying this aggregation process is crucial for optimizing multi-agent systems' performance. Mechanisms that foster collaboration among agents are vital to prevent locally optimal decisions from leading to globally suboptimal results.

\begin{table*}[t]
\centering
\caption{Comparison of Representative Surveys alligned with Our Contribution}

\label{tab:rw_contribution_comparison}
\scriptsize 
\renewcommand{\arraystretch}{1.25}
\begin{tabular}{| p{1cm} | p{2.0cm} | p{4.2cm} | p{8.2cm} |}
\hline
\textbf{Citation} & \textbf{Scope / Domain} & \textbf{Key Insights from Prior Work} & \textbf{Complementary Contribution of Our Work} \\

\hline
\hline

\cite{b2073} & DRL-driven offloading in aerial access networks &
Explores DRL-based computational task offloading for UAV and aerial platforms, emphasizing mobility patterns, aerial coverage, and DRL decision policies. &
Our work focuses on ground Vehicular Edge Computing (VEC) and extends the discussion to road-side MEC, vehicular mobility, and multi-tier architectures. We contextualize DRL methods within V2X dynamics, deadline constraints, and architecture-aware offloading strategies. \\
\hline
\cite{b2074} & DRL survey for IoT &
Provides an extensive overview of DRL algorithms and IoT applications, covering model-free and model-based methods, exploration strategies, and training challenges. &
We bring these algorithmic foundations into the VEC setting by analyzing how DRL interacts with vehicular mobility, task deadlines, and fluctuating communication environments. We further relate DRL design to specific offloading architectures (centralized, distributed, hierarchical). \\
\hline
\cite{b175} & MARL for vehicular networks &
Discusses MARL algorithms and vehicular networking coordination, focusing on communication strategies, cooperation, and joint decision-making. &
Our work complements this by linking MARL formulations directly to vehicular computational task offloading workflows, including POMDP-to-MDP modeling, reward design (latency-energy trade-offs), and coordination among heterogeneous vehicular agents. \\
\hline
\cite{b176} & DRL for edge caching &
Highlights DRL use in mobile edge caching, user content prediction, and proactive resource placement. &
We address a different layer of MEC functionality, computation offloading. Our work integrates DRL/MARL approaches for offloading decisions, vehicle-induced non-stationarity, and architecture-oriented learning strategies, which are not the focus of caching studies. \\
\hline
\cite{b174}  & MARL for future Internet &
Provides a comprehensive view of MARL algorithms, including Centralized Training and Decentralized Execution (CTDE), value decomposition, and communication-efficient learning. &
Our work builds upon these algorithmic insights by examining their applicability to VEC offloading. We relate MARL mechanisms to vehicular task management, shared rewards, multi-agent synchronization and generalized learning in real-time adaptation. \\
\hline
\cite{b203} & RL/DRL for vehicular computational task offloading &
Reviews RL/DRL approaches for vehicular offloading and vehicular cloudlets, discussing energy–latency trade-offs. &
We extend this domain-focused perspective by offering a unified architectural taxonomy (centralized, distributed, hierarchical), enhanced MDP/POMDP formulation analysis, and a deeper examination of DRL coordination and generalization challenges under high mobility and dynamic V2X conditions. \\

\hline 

\end{tabular}
\end{table*}

\subsection{Our Contribution}
This survey systematically investigates how DRL can be applied to optimize computational task offloading in VEC across three primary architectural paradigms:  
(i) \textbf{Centralized Offloading and Computing}, where a central server manages offloading decisions using a single DRL agent with global state awareness;  
(ii) \textbf{Distributed Offloading and Computing}, where multiple edge nodes coordinate through DRL to balance computation loads and handle decentralized service delivery; and  
(iii) \textbf{Hierarchical Offloading and Computing}, where heterogeneous layers, including vehicles, edge servers, UAVs, and cloud platforms, collaborate to execute tasks efficiently in dynamic, multi-tier environments.

The core objective of this survey is to analyze how MDP formulations and DRL methods are deployed across these paradigms, particularly under uncertainty and dynamic conditions. We highlight critical technical challenges such as value function approximation, policy convergence, action space design (discrete vs. continuous), exploration-exploitation trade-offs, and the role of centralized versus decentralized learning in multi-agent systems. Special attention is given to the design of reward functions and the coordination of partially observable agents to ensure both local and global system performance.

The key contributions of this survey are as follows:
\begin{itemize}
    \item We provide a structured taxonomy of DRL-based computational task offloading solutions categorized by centralized, distributed, and hierarchical architectures, illustrating how DRL techniques align with each system's operational characteristics and resource constraints.
       
    \item We assess the variability and limitations of reward function design, particularly in multi-objective settings where latency, energy, and fairness must be jointly optimized, an aspect often neglected or improperly balanced in existing studies.   

    \item We examine the role of action space representation (continuous vs. discrete), policy types (stochastic vs. deterministic), and value approximation strategies in shaping the scalability and convergence of DRL algorithms in both single-agent and multi-agent settings.
    
    \item We explore multi-agent learning frameworks, including CTDE, and analyze the challenges of coordination, synchronization, and uniform learning in heterogeneous agent environments.
    
    \item We identify open research directions around scalable DRL design, dynamic coordination, and hierarchical multi-agent decision-making in VEC, offering guidance for future work on robust and adaptive DRL frameworks in real-world vehicular networks.
\end{itemize}

\subsection{Organization of the Studies}

This survey is organized to reflect the layered architectural applications of DRL in VEC. 
Section~\ref{sec:computing_paradigm} outlines the core computing paradigms in VEC, while Section~\ref{sec:network_topology} introduces common network architectures, namely, centralized, distributed, and hierarchical, and their relevance to dynamic vehicular environments. Section~\ref{sec:drl_fundamentals} presents foundational DRL techniques, including value-based, policy-based, and actor-critic methods, as well as multi-agent learning frameworks applicable to edge computing. Section~\ref{sec:single_server} focuses on centralized offloading models in which a single agent manages computation and resource allocation. Section~\ref{sec:multi_server} reviews distributed DRL models involving coordination among multiple edge servers through decentralized or federated learning. Section~\ref{sec:hierarchical} explores heterogeneous, multi-tier offloading systems that incorporate vehicles, edge servers, UAVs, and cloud nodes to optimize decision-making in layered infrastructures. Each of these sections examines DRL methods, optimization objectives, reward designs, and system limitations under both single-agent and multi-agent settings. Section~\ref{sec:lessons} synthesizes the key lessons learnt and open research directions for advancing DRL-based offloading in VEC systems, and Section~\ref{sec:conclusion} concludes with a summary of findings.

\section{Computing Paradigms in Vehicular Edge Networks}
\label{sec:computing_paradigm}
The increasing computational and communication demands of ITS, such as autonomous driving, real-time traffic monitoring, and vehicle-to-everything (V2X) applications, have necessitated the development of advanced computing paradigms capable of supporting delay-sensitive and resource-intensive operations. These paradigms enable vehicles to offload computational workloads to external resources, compensating for the inherent limitations of onboard hardware in terms of processing power, energy capacity, and storage. This section categorizes the major offloading paradigms in VEC into three groups: centralized, edge-centric, and mobile/ubiquitous architectures. We discuss their operational characteristics, design trade-offs, and deployment challenges in the context of vehicular environments.

\subsection{Centralized Paradigm}
Centralized cloud computing (CC) provides high-throughput, flexible computing resources hosted in remote data centers~\cite{b101}. These cloud platforms support large-scale processing and data aggregation, making them ideal to deliver scalable, cost-effective, and high-performance computation. Vehicles connect to cloud infrastructure through cellular or broadband networks, uploading raw or preprocessed data for centralized computation. Despite their computational abundance, centralized clouds face significant latency and reliability challenges due to their physical separation from mobile users~\cite{b102}. The performance of cloud-based systems degrades for applications requiring stringent QoS guarantees, particularly in highly dynamic vehicular contexts where service continuity and fast reaction times are crucial. Additionally, the high volume of data transmission can saturate backhaul links, leading to network congestion and increased service cost.

\subsection{Edge-Centric Paradigms}
To overcome the latency bottlenecks of cloud computing, edge-centric paradigms aim to localize computation by deploying infrastructure closer to end-users.

\subsubsection{Mobile Edge Computing (MEC)}
MEC~\cite{b104} integrates compute and storage capabilities into the cellular network infrastructure, typically co-located with base stations, RSUs, or micro data centers. MEC nodes serve as intermediary layers that reduce the latency between vehicles and remote services, allowing real-time decisions in safety-critical scenarios such as collision avoidance or traffic light optimization. MEC also allows mobile network operators to expose their Radio Access Networks (RAN) to third-party application developers, enabling deployment of customized vehicular services with direct access to user location, mobility patterns, and radio conditions. However, effective MEC deployment is non-trivial. It requires optimal server placement that accounts for vehicular mobility patterns, task heterogeneity, and spatiotemporal demand variation~\cite{b202}. Static spectrum and resource allocation schemes often lead to underutilization or congestion, necessitating adaptive and context-aware resource orchestration.

\subsubsection{Fog Computing (FC)}
Fog computing~\cite{b103} complements MEC by introducing an additional intermediate layer of distributed resources that exist between the edge and centralized cloud. Fog nodes are often deployed in fixed infrastructure such as traffic lights, parking stations, or gateway routers. These nodes extend computing and storage services further into the access network, offering more localized processing with enhanced resilience to core network failures. Unlike MEC, which typically focuses on cellular RAN integration, fog computing emphasizes decentralized coordination among heterogeneous devices and supports horizontal offloading between neighboring nodes. This is especially useful in distributed vehicular systems where multi-hop communication and regional collaboration are essential.

\subsection{Ubiquitous and Mobile Paradigms}
Mobile paradigms leverage vehicular and aerial platforms as computing nodes. These systems offer high flexibility and dynamic service provisioning, particularly in infrastructure-sparse or delay-critical environments.

\subsubsection{Vehicular Cloudlets/Onboard Servers}
Vehicular Cloudlets or Onboard Vehicle Servers (OVS)~\cite{b105} form decentralized micro-clouds using the spare computational, storage, and communication resources of vehicles. Vehicles equipped with sufficient onboard capacity participate in cooperative computing by sharing tasks with neighboring nodes through V2V or V2I links. This opportunistic paradigm reduces the dependency on fixed infrastructure and supports localized processing of time-sensitive data. Vehicular cloudlets are particularly well-suited for collaborative perception, platoon coordination, and distributed environment sensing. However, they face several challenges, including network volatility due to vehicle mobility, uncertain node availability, security threats, and the need for lightweight consensus mechanisms to manage distributed decision-making.

\subsubsection{Unmanned Aerial Vehicles (UAVs)}
UAVs ~\cite{b1061}, or drones, are emerging as agile computing platforms that offer vertical mobility and line-of-sight communication. UAVs can be equipped with edge servers to provide coverage and computational assistance in dynamic, high-density, or disaster-prone regions where terrestrial infrastructure is unavailable or insufficient. In ITS, UAVs can relay data from roadside sensors, support emergency communication, assist in coverage extension for V2X links, and serve as mobile task execution platforms. Their integration into vehicular networks introduces new degrees of freedom for network planning and service distribution. Nevertheless, UAV-based systems must address constraints such as limited battery life, collision avoidance, flight path optimization, and regulatory compliance.

\subsection{Comparative Summary}
Table~\ref{tab:computing_paradigms_comparison} summarizes the primary trade-offs among the discussed computing paradigms in terms of key performance criteria relevant to VEC.

\begin{table*}[!t]
\centering
\caption{Comparison of Vehicular Computing Paradigms}
\label{tab:computing_paradigms_comparison}
\begin{tabular}{|c|c|c|c|c|c|}
\hline
\textbf{Paradigm} & \textbf{Latency} & \makecell[c]{\textbf{Mobility}\\\textbf{Support}} & \makecell[c]{\textbf{Deployment}\\\textbf{Cost}} & \textbf{Scalability} & \textbf{Reliability} \\
\hline
\hline
CC & High & Low & Low & High & Medium \\
\hline
MEC & Low & Medium & Medium & Medium & High \\
\hline
FC & Medium & Medium & Medium & Medium & High \\
\hline
OVS & Very Low & High & Low & Medium & Low-Medium \\
\hline
UAV & Low & Very High & High & Medium & Medium \\
\hline
\end{tabular}
\vspace{-0.75em}
\end{table*}

\section{Network Architectures in VEC}
\label{sec:network_topology}
Vehicular networks demand low-latency communication, high mobility support, and dynamic resource allocation. As computation offloading becomes essential for connected and autonomous vehicles, the design of underlying network architectures plays a critical role in system efficiency. This section classifies VEC topologies into three primary models: centralized, distributed, and hierarchical. We discuss their structural characteristics, deployment implications, and operational challenges. Furthermore, we examine ad-hoc and heterogeneous subtypes that extend the capabilities of hierarchical architectures for more flexible and resilient computation.

\subsection{Centralized MEC Architecture}
In a centralized MEC setup~\cite{b202}, a single MEC server, which is typically co-located at a macro base station (MBS) and manages all computation offloading from nearby vehicles. This model benefits from centralized control over radio access resources, simplified policy enforcement, and efficient coordination using channel state information (CSI) and aggregated task requests. By processing tasks closer to the user than traditional cloud infrastructure, centralized MEC significantly reduces communication latency. However, the reliance on a single server introduces scalability bottlenecks. As traffic load increases, the limited computational and bandwidth resources at the MEC node can cause task queuing, degraded latency, or packet loss. Moreover, this architecture is vulnerable to single points of failure and may not scale well in high-density vehicular environments.

\subsection{Distributed MEC Architecture}
Distributed MEC~\cite{b150} addresses the limitations of centralized models by deploying multiple edge nodes (e.g., RSUs, vehicles, or microservers) throughout the network. These nodes collaboratively manage computation, storage, and control tasks. Processing data near its source allows real-time responses, especially in latency-critical applications like collision avoidance or traffic signal optimization. By distributing the computational workload, this architecture improves scalability, adaptability, and fault tolerance. Additionally, local task processing enhances data privacy, as sensitive information need not traverse wide-area networks. Nevertheless, distributed systems incur increased coordination overhead and require efficient task allocation mechanisms to avoid resource underutilization or duplication of effort.

\subsection{Hierarchical MEC Architecture}
Hierarchical edge computing~\cite{b151, b152} introduces a multi-tiered resource structure, integrating local, regional, and cloud layers. Tasks are dynamically mapped to the most appropriate tier based on urgency, complexity, and resource availability. For example, simple control tasks may be executed at the vehicle or RSU level, while compute-intensive analytics can be offloaded to regional fog nodes or centralized clouds. Technologies such as SDN, NFV, and network slicing enable flexible control, service orchestration, and dynamic scaling across layers. The hierarchical model supports effective load balancing and fault tolerance, making it particularly suitable for large-scale, heterogeneous deployments such as smart cities.

\subsubsection{Ad-hoc Networks in MEC}
Vehicular Ad-hoc Networks (VANETs)~\cite{b153, b154} and UAV-assisted links~\cite{b155, b156} constitute decentralized, infrastructure-independent topologies. Vehicles, drones, or other mobile nodes form dynamic peer-to-peer networks that support local offloading and communication. Such ad-hoc topologies are highly responsive and resilient in scenarios lacking fixed infrastructure, such as disaster recovery, remote highways, or temporary traffic congestion zones. Direct communication among nodes (e.g., vehicles and UAVs or UAVs) reduces dependency on base stations and mitigates backhaul congestion. However, ensuring stable links and reliable handoffs in mobile and unpredictable topologies remains a key challenge.

\subsubsection{Heterogeneous MEC}
Heterogeneous MEC refers to integrated topologies involving diverse network elements, such as, vehicles, RSUs, satellites, drones, and clouds, coexisting to provide flexible offloading solutions~\cite{b205}. These architectures enable seamless transitions between communication modes and computational domains, ensuring coverage across urban, rural, and remote areas. This model is particularly beneficial for long-range vehicular communication and applications requiring global coordination. For instance, satellite-assisted MEC supports data relay in infrastructure-sparse regions, while terrestrial fog and edge nodes handle high-throughput tasks in urban corridors. The key challenge lies in harmonizing resource heterogeneity and maintaining interoperability across protocols and hardware platforms.

\subsection{Comparative Summary}
Table~\ref{tab:network_topologies_comparison} provides a comparative view of the discussed topologies with respect to their scalability, latency performance, mobility support, control complexity, and overhead.

\begin{table*}[!t]
\centering
\caption{Comparison of Vehicular Edge Network Topologies}
\label{tab:network_topologies_comparison}
\begin{tabular}{|c|c|c|c|c|c|}
  \hline
  \textbf{Topology} & \textbf{Latency} & \makecell[c]{\textbf{Mobility}\\\textbf{Support}} & \textbf{Scalability} & \makecell[c]{\textbf{Control}\\\textbf{Complexity}} & \makecell[c]{\textbf{Coordination}\\\textbf{Overhead}} \\
\hline
\hline
Centralized MEC & Low (under load: high) & Medium & Low & Low & Low \\
\hline
Distributed MEC & Very Low & High & High & Medium & High \\
\hline
Hierarchical MEC & Variable (tier-dependent) & High & High & High & Medium \\
\hline
Ad-hoc MEC & Very Low & Very High & Medium & Very Low & High \\
\hline
Heterogeneous MEC & Low-Medium & Very High & Very High & High & Very High \\
\hline
\end{tabular}
\vspace{-0.75em}
\end{table*}



\section{Fundamentals of DRL}
\label{sec:drl_fundamentals}
This section provides an overview of foundational principles in DRL, beginning with the MDP, followed by value function approximation, prominent DRL algorithms, and their extensions to multi-agent settings relevant to VEC.

\subsection{Markov Decision Process}
RL models sequential decision-making problems in which an agent interacts with an environment by observing states, selecting actions, and receiving rewards, with the goal of maximizing long-term cumulative return (Fig.~\ref{fig:rl_mdp}). This interaction is commonly formulated as a Markov Decision Process (MDP) defined by the tuple $(\mathcal{S}, \mathcal{A}, P, R, \gamma)$, where $\mathcal{S}$ and $\mathcal{A}$ denote the state and action spaces, $P$ the state transition dynamics, $R$ the reward function, and $\gamma$ the discount factor~\cite{b131}. Under the Markov property, transitions depend only on the current state and action. RL policies may be deterministic or stochastic, and optimal decision-making is characterized through value and action-value functions satisfying the Bellman optimality conditions.

\begin{figure}[!t]
\centering
\begin{tikzpicture}[
    >=Latex,
    node distance=2cm and 2.5cm,
    every node/.style={font=\small},
    box/.style={draw, rounded corners, minimum width=2.5cm, minimum height=1.2cm, align=center}
]

\node[box, fill=gray!15] (agent) {Agent\\- Learned Policy (Exploitation)\\- Random Policy (Exploration)};
\node[box, fill=orange!20, below=of agent] (env) {\textbf{Environment}};
\node[box, fill=yellow!25, right=1.5cm of agent] (buffer) {Replay Buffer\\$(s_t, a_t, r_t, s_t')$};

\draw[->, thick, red] (agent.south) -- node[right=2pt] {action $a_t$} (env.north);
\draw[->, thick, blue] (env.west) -- ++(-1.5,0) node[midway, above] {state $s_t$} |- (agent.west);
\draw[->, thick, black] (env.north east) to[out=45,in=-45] node[right=2pt] {reward $r_t$} (agent.south east);
\draw[->, thick, green!60!black] (env.north west) to[out=135,in=-135] node[right=2pt] {next state $s_{t}'$} (agent.south west);

\draw[->, thick, orange] (buffer.west) -- (agent.east) node[midway, below] {mini-batch};

\draw[->, thick, violet] (agent.north) -- ++(0,0.7) -| node[pos=0.25, below] {store $(s_t, a_t, r_t, s_t')$} (buffer.north);

\end{tikzpicture}
\caption{Agent-environment interaction loop in RL.}
\vspace{-0.75em}
\label{fig:rl_mdp}
\end{figure}

\subsection{Value Function Approximation via Value Iteration}
Value function approximation replaces tabular value representations with parameterized function approximators, typically neural networks, and enables RL to scale to high-dimensional or continuous state spaces~\cite{b134}. Instead of storing explicit value tables, the agent learns to estimate state-value or action-value functions by minimizing the temporal-difference (TD) error between predicted and target values. For action-value approximation, the learning objective can be expressed as
\begin{align}
L(\theta) = \mathbb{E}_{(s,a,r,s')} \Big[ \big( r + \gamma \max_{a'} \hat{Q}(s',a';\theta) - \hat{Q}(s,a;\theta) \big)^2 \Big],
\end{align}
where $\theta$ denotes the parameters of the function approximator. This formulation forms the basis of deep Q-learning and related action value based evaluation frameworks, which estimate the long-term utility of state-action pairs and are widely used in vehicular edge computing to address complex environment dynamics.
\subsubsection{Deep Q-Network (DQN) and Variants}
The DQN~\cite{b135} is an off-policy RL algorithm that learns the optimal policy from transitions stored in an experience buffer, which are collected using an earlier version of the policy. In DQN, a neural network is used to estimate the current state-action values (Q-values). To stabilize training, a target network is maintained, which is updated less frequently and used to compute target Q-values. This helps reduce the instability caused by correlations in sequential training data. Double DQN~\cite{b136} addresses the overestimation bias in standard DQN by using the main network to select actions and the target network to evaluate them. This refinement produces more accurate Q-value estimates. Dueling DQN further improves performance by separately estimating the state-value and advantage functions. These are then combined to produce more robust action-value predictions, enhancing generalization and learning efficiency, particularly in environments where certain actions have little effect on the outcome.

\subsection{Value Function Approximation via Policy Gradient}
Policy gradient methods optimize the policy directly by estimating the gradient of the expected return with respect to the policy parameters. The gradient of the value function is expressed as the expectation of the trajectory return multiplied by the score function~\cite{b131}:
\begin{align}
\nabla_{\theta} V^\pi(s_0) &\propto \sum_{s \in \mathcal{S}} d^\pi(s) \sum_{a \in \mathcal{A}} Q^\pi(s, a) \nabla_{\theta} \pi(a \mid s) \notag \\
&= \sum_{s \in \mathcal{S}} d^\pi(s) \sum_{a \in \mathcal{A}} \pi(a \mid s) Q^\pi(s, a) \nabla_{\theta} \log \pi(a \mid s) \notag \\
&= \mathbb{E}_{\pi} \left[ Q^\pi(S_t, A_t) \nabla_{\theta} \log \pi_{\theta}(A_t \mid S_t) \right]
\end{align}

\subsubsection{Actor-Critic Method}
The Actor-Critic method \cite{b139} consists of two components: the \textit{Actor}, which selects actions based on the current state (i.e., the policy), and the \textit{Critic}, which estimates the value of those actions (i.e., the action-value function $Q(s, a)$). The Critic updates its value estimates by minimizing the temporal-difference (TD) error based on the Bellman equation. The Actor updates its policy by ascending the gradient of the expected return, guided by feedback from the Critic. This interaction allows the agent to learn more effectively from continuous environments and improves the stability of policy updates.

\subsubsection{Trust Region Policy Optimization (TRPO)}
TRPO~\cite{b142} is an on-policy Actor-Critic algorithm that constrains the magnitude of policy updates to maintain stability. Policy updates are computed using trajectories generated by the current policy and are limited by a trust region defined using the Kullback-Leibler (KL) divergence between the old and new policies. This constraint ensures that each policy update does not deviate too far from the previous policy, promoting more consistent and reliable improvement over time.

\subsubsection{Proximal Policy Optimization (PPO)}
PPO~\cite{b143} is another on-policy Actor-Critic method that improves training stability using a clipped surrogate objective. PPO avoids the complexity of constrained optimization in TRPO by penalizing policy updates that go beyond a predefined threshold. Additionally, PPO includes an entropy term in the loss function to encourage exploration by maintaining stochasticity in the policy. This balance between exploration and exploitation leads to more sample-efficient and stable learning.

\subsubsection{Soft Actor-Critic (SAC)}
SAC~\cite{b140} is an off-policy algorithm that enhances learning efficiency in continuous control environments by maximizing a stochastic policy augmented with an entropy term. The algorithm consists of three components: a policy network, two Q-networks, and a value network. The policy network generates actions that maximize both expected return and entropy, promoting diverse exploration. The Q-networks estimate expected return, while the value network predicts the state value to stabilize training by minimizing the TD error. SAC effectively balances exploration and exploitation, making it well-suited for complex tasks.

\subsubsection{Deep Deterministic Policy Gradient (DDPG)}
The DDPG~\cite{b141} algorithm is designed for continuous action spaces. It employs an Actor network to produce deterministic actions based on the current state and a Critic network to estimate the corresponding Q-values. The Critic is trained by minimizing the TD error, and the Actor is updated by ascending the gradient of the Q-value with respect to the policy parameters. To improve stability, DDPG uses soft target networks for both Actor and Critic, which are updated gradually. DDPG enables stable learning in continuous domains that require precise action control.

\subsubsection{Twin Delayed Deep Deterministic Policy Gradient (TD3)}
The TD3~\cite{b157} improves upon DDPG by addressing overestimation bias and training instability. TD3 employs two Critic networks and selects the minimum predicted Q-value to reduce overestimation. Additionally, it delays Actor updates relative to Critic updates and introduces target policy smoothing by adding noise to the target actions. These enhancements lead to more robust and stable learning in continuous action spaces, improving performance in complex control environments.

Table~\ref{tab:drl_comparison} compares key DRL algorithms in terms of core characteristics, training paradigms, features, and computational complexity.

\begin{table*}[!t]
\centering

\caption{Comparison of Key DRL Algorithms}
\label{tab:drl_comparison}
\begin{tabular}{|
c|c|c|>{\centering}m{1.5cm}|c|
>{\color{red}\tiny}m{1.5cm}|
m{5.5cm}|
} \hline
\textbf{Algorithm} & \textbf{Policy} & \textbf{Learning} & \textbf{Action Space} & \makecell[c]{\textbf{Update}\\\textbf{Type}} & \footnotesize \textbf{Complexity} & \textbf{Key Features and Enhancements} \\
\hline
\hline
DQN & Deterministic & Off-policy & Discrete & Value-based & $\mathcal{O}(N_{Q})$ & Uses replay buffer and target networks to stabilize Q-learning \\
\hline
Double DQN & Deterministic & Off-policy & Discrete & Value-based & $\mathcal{O}(N_{Q})$ & Reduces Q-value overestimation by decoupling action selection and evaluation \\
\hline
Dueling DQN & Deterministic & Off-policy & Discrete & Value-based & $\mathcal{O}(N_{Q})$ & Separates state-value and advantage estimations for more robust learning \\
\hline
DDPG & Deterministic & Off-policy & Continuous & Actor-Critic & $\mathcal{O}(N_{actor}+N_{critic})$ & Learns a deterministic policy with soft target updates for stable training \\
\hline
TD3 & Deterministic & Off-policy & Continuous & Actor-Critic & $\mathcal{O}(N_{actor}+2N_{critic})$ & Improves DDPG with twin critics, delayed actor updates, and target smoothing \\
\hline
PPO & Stochastic & On-policy & Discrete \& Continuous & Actor-Critic & $\mathcal{O}(K \cdot N\times(N_{actor}+N_{critic}))$ & Uses clipped policy objective and entropy bonus for stable and efficient updates \\
\hline
TRPO & Stochastic & On-policy & Discrete \& Continuous & Actor-Critic & $\mathcal{O}(K \cdot N\times(I\cdot N_{actor}+N_{critic}))$ & Constrains policy updates using trust regions based on KL divergence \\
\hline
SAC & Stochastic & Off-policy & Continuous & Actor-Critic & $\mathcal{O}(N_{actor}+2N_{critic})$ & Optimizes stochastic policy with entropy maximization for robustness \\
\hline
\end{tabular}

\vspace{0.5em}
{\footnotesize \textit{Note:}$N_{Q}=\text{Number of Q Network  Parameter}$, , $N=\text{Number of Minibatch}$, $K=\text{Number of Epoch}$,$I=\text{Number of epoch for Conjugate Gradient}$,  $N_{actor}=\text{Number of Actor Network  Parameter}$,$N_{critic}=\text{Number of Critic Network Parameter}$ }
\end{table*}

\subsection{Multi-Agent RL (MARL)}
In VEC, system-level objectives such as latency reduction, energy efficiency, load balancing, scalability, reliability, and data security can be addressed using either a centralized single-agent framework modeled as a fully observable MDP, or a distributed multi-agent framework modeled as a POMDP. In single-agent RL, a central controller may optimize latency, energy consumption, and task assignment across vehicles and servers. In contrast, MARL~\cite{b146, b147} enables multiple autonomous agents to interact with a shared environment, each optimizing its own objective while contributing to overall system performance.

MARL paradigms can be categorized based on the nature of agent interaction:
\begin{itemize}
    \item Collaborative MARL: All agents work toward a common goal, such as minimizing aggregate latency or energy consumption, without focusing on individual benefits~\cite{b162}.
    \item Cooperative MARL: Agents optimize their individual objectives but cooperate when beneficial, for instance via joint spectrum sensing or traffic coordination~\cite{b163}.
    \item Competitive MARL: Agents pursue conflicting objectives and compete for shared resources like channels or bandwidth~\cite{b164}.
\end{itemize}

\subsubsection{Key Challenges in MARL}
Multi-agent systems introduce several challenges not present in single-agent RL. First, agents often operate under partial observability, relying only on local information. This leads to the non-stationarity problem~\cite{b165}, where concurrently learning agents alter the environment dynamics for each other, violating the stationarity assumptions required for convergence. Second, scalability becomes a significant concern~\cite{b166}; as the number of agents increases, the joint action space grows exponentially, increasing the computational complexity of finding optimal policies. While deep neural networks (DNNs) have been adopted in MARL to approximate value functions and policies in large spaces, they introduce their own challenges in terms of convergence, stability, and interpretability. Consequently, MARL algorithms must manage a trade-off between performance, coordination overhead, and scalability in highly dynamic vehicular environments.

\subsubsection{CTDE}
The CTDE paradigm~\cite{b147a,b167} is widely adopted in MARL to address coordination and non-stationarity challenges. Under CTDE, agents are trained using centralized information, where a centralized critic or learning mechanism has access to the global state and joint action information of all agents. This joint perspective enables the optimization of individual agent policies in a coordinated manner while explicitly accounting for inter-agent interactions. During execution, however, agents operate independently using only their local observations and limited communication, without access to global information. This separation between training and execution reduces communication overhead, improves scalability, and enables decentralized real-time decision-making.

During the training phase, centralized access helps mitigate two fundamental challenges in MARL: non-stationarity, arising from simultaneous policy updates by multiple agents, and partial observability, where local observations alone are insufficient for optimal decision-making. By leveraging centralized critics, shared gradients, or inter-agent backpropagation, CTDE exploits global state and joint action information to stabilize learning dynamics and improve credit assignment. By explicitly modeling the evolving behaviors of other agents during training, CTDE promotes coordinated policy learning and enhances robustness in dynamic multi-agent environments. Consequently, CTDE has become a foundational framework for many state-of-the-art MARL algorithms, including centralized-critic and value-decomposition approaches.

\begin{itemize}[wide=0pt]
    \item \textbf{Multi-Agent DDPG (MADDPG)}~\cite{b144}: MADDPG extends the DDPG framework to multi-agent systems. Each agent is equipped with a local actor and a centralized critic that accesses the global state and joint actions during training. This enables more accurate value estimation in complex multi-agent interactions that involve both cooperation and competition. At execution time, each agent uses only its actor and local observations, enabling decentralized and autonomous behavior. While MADDPG is effective in modeling multi-agent dynamics, its dependence on full global observability during training limits its scalability. Additionally, it lacks explicit mechanisms for reward attribution or value decomposition, making it difficult to quantify individual agent contributions to team performance.

    \item \textbf{Counterfactual Multi-Agent Policy Gradient (COMA)}~\cite{b169}: COMA builds on the actor-critic architecture to address the credit assignment problem in cooperative environments. Unlike MADDPG, COMA uses a centralized critic that computes a counterfactual baseline to evaluate the individual contribution of each agent to the global reward. This baseline compares the actual joint action with an alternative where one agent takes a different action while others remain fixed. This improves credit assignment and encourages more efficient policy updates. COMA is well-suited for cooperative tasks but assumes access to the full global state, which may be impractical in large-scale, partially observable settings.

    \item \textbf{Value Decomposition Network (VDN)}~\cite{b172}: VDN simplifies learning in cooperative MARL by decomposing the global Q-function into a sum of individual agent Q-functions. Each agent learns to maximize its local Q-value, which contributes additively to the team objective. This decomposition enables decentralized execution while maintaining alignment with a shared reward signal. VDN is particularly effective in structured cooperative settings, where individual agent contributions are approximately independent. However, the assumption of additivity may fail to capture complex interdependencies among agents in more intricate environments.
\end{itemize}

\subsubsection{Decentralized Learning with Networked Agents}
Decentralized learning~\cite{b168} is particularly effective in large-scale multi-agent systems, such as IoT or vehicular networks, where centralized coordination becomes impractical. In this paradigm, each agent relies primarily on its own local observations and, in some cases, partial information from neighboring agents. This minimizes the need for global state information, thereby reducing both communication overhead and computational burden. The decentralized structure offers high scalability and robustness, making it well-suited for dynamic environments. However, the lack of centralized oversight introduces challenges related to \textbf{non-stationarity}. As each agent learns and adapts independently, the environment becomes unstable from the perspective of any individual agent. This can lead to policy divergence and suboptimal convergence in the absence of effective coordination mechanisms.

Several algorithms have been developed to address decentralized learning in MARL:
\begin{itemize}[wide=0pt]
    \item \textbf{Multi-Agent Deep Q-Network (MADQN)}~\cite{b170}: In MADQN, each agent independently learns a Q-function using local observations, enabling autonomous decision-making without direct inter-agent communication. A global reward signal is shared across agents to promote cooperative behavior. MADQN is well-suited for resource-constrained environments such as V2X communications, where agents must manage heterogeneous and dynamic traffic. To improve scalability, the action space can be reduced using techniques like virtual agents, which enhance learning efficiency under varying demand conditions.

    \item \textbf{Multi-Agent Actor-Critic (MAAC)}~\cite{b168}: MAAC extends the actor-critic framework to decentralized settings. Each agent is equipped with a local actor and critic, allowing it to evaluate and update its policy based on its own observations. In some implementations, agents exchange partial estimates of the global action-value function with nearby peers, enabling limited coordination. This method balances the benefits of decentralized execution with collaborative learning, making it effective in partially observable, multi-agent environments.

    \item \textbf{Multi-Agent Soft Actor-Critic (MASAC)}~\cite{b59}: MASAC adapts the SAC algorithm to decentralized scenarios. Each agent optimizes its own stochastic policy using entropy-augmented objectives, encouraging exploration while learning to act optimally. MASAC does not require direct inter-agent communication, making it especially valuable in environments where communication is limited, unreliable, or expensive. Its entropy-regularized objective helps maintain robustness under uncertainty and fosters diversity in agent behavior.
\end{itemize}

Table~\ref{tab:marl_comparison} summarizes key characteristics of representative MARL algorithms, including their policy types, training and execution paradigms, use of centralized critics, communication requirements, associated computational complexity, and suitability for VEC applications.

\begin{table*}[!t]
\centering
\caption{Comparison of Key MARL Algorithms}
\label{tab:marl_comparison}
\begin{tabular}{|c|c|c|c|c|c|>{\color{red}\tiny}m{1.5cm}|m{3.5cm}|}
  \hline
\textbf{Algorithm} & \textbf{Policy} & \textbf{Training} & \textbf{Execution} &
\makecell[c]{\textbf{Central}\\\textbf{Critic}} & \textbf{Communication} &
\footnotesize \textbf{Complexity} & \textbf{Notes} \\
\hline
\hline

MADDPG & Deterministic & Centralized & Decentralized & Yes & No &
$\mathcal{O}\big(N\times N_{\text{actor}} + N^{2} N_{\text{critic}}\big)$ &
Suitable for mixed cooperative and competitive tasks; suffers from scalability and lacks explicit credit assignment; uses centralized critic per agent \\
\hline

COMA & Stochastic & Centralized & Decentralized & \makecell{Yes\\(counterfactual)} & No &
$\mathcal{O}\big(N\,N_{\text{critic}}\big)$ &
Effective for credit assignment in cooperative tasks; uses a single central critic \\
\hline

VDN & Deterministic & Centralized & Decentralized &
\makecell{Implicit\\(decomposition)} & No &
$\mathcal{O}\big(N\,N_Q\big)$ &
Assumes additive Q-function; scalable but limited in capturing inter-agent dependencies \\
\hline

MADQN & Deterministic & Decentralized & Decentralized &
No & No &
$\mathcal{O}\big(N\,N_Q\big)$ &
Each agent uses local Q-function and global reward; effective for V2X and resource-constrained systems \\
\hline

MAAC & Stochastic & Decentralized & Decentralized &
Local & \makecell{Partial\\(neighbor-based)} &
$\mathcal{O}\big(N\,N_{\text{b}}\,N_{\text{critic}}\big)$ &
Scalable; enables limited collaboration through local communication and shared estimates \\
\hline

MASAC & Stochastic & Decentralized & Decentralized &
No & No &
$\mathcal{O}\big(N\,(2N_{\text{critic}} + N_{\text{actor}})\big)$ &
Applies entropy regularization for robust exploration; suitable when communication is limited \\
\hline

\end{tabular}

\vspace{0.5em}
{\footnotesize \textit{Note:} $N=\text{Number of Agents}$, $N_b=\text{Number of Neigbour}$,$N_{actor}=\text{Number of Actor Network  Parameter}$,$N_{Q}=\text{Number of Q Network  Parameter}$, $N_{critic}=\text{Number of Critic Network Parameter}$ }
\end{table*}

\subsubsection{Game Theory Integrated MARL}
Game theory and MARL~\cite{b179, b180} provide complementary perspectives for modeling decision-making in dynamic, multi-agent environments. Game theory offers strategic frameworks such as Nash equilibrium and Stackelberg equilibrium, which formalize the interactions among rational agents by defining stable outcomes where no agent can unilaterally improve its utility. In contrast, MARL emphasizes learning optimal policies through trial-and-error interactions with the environment, aiming to maximize cumulative rewards over time. Unlike game-theoretic models, MARL typically does not assume full rationality or complete knowledge of other agents' strategies. Integrating game-theoretic concepts into MARL introduces a layer of strategic reasoning that enhances coordination and policy convergence. This hybrid approach enables agents to learn equilibrium-aware behaviors, effectively balancing cooperation and competition. As a result, game-theoretic MARL facilitates more stable learning dynamics and improved performance in decentralized systems where agents must adapt to the behaviors of others in real time.

\begin{figure}[!t]
  \centering
    \includegraphics[width=\linewidth]{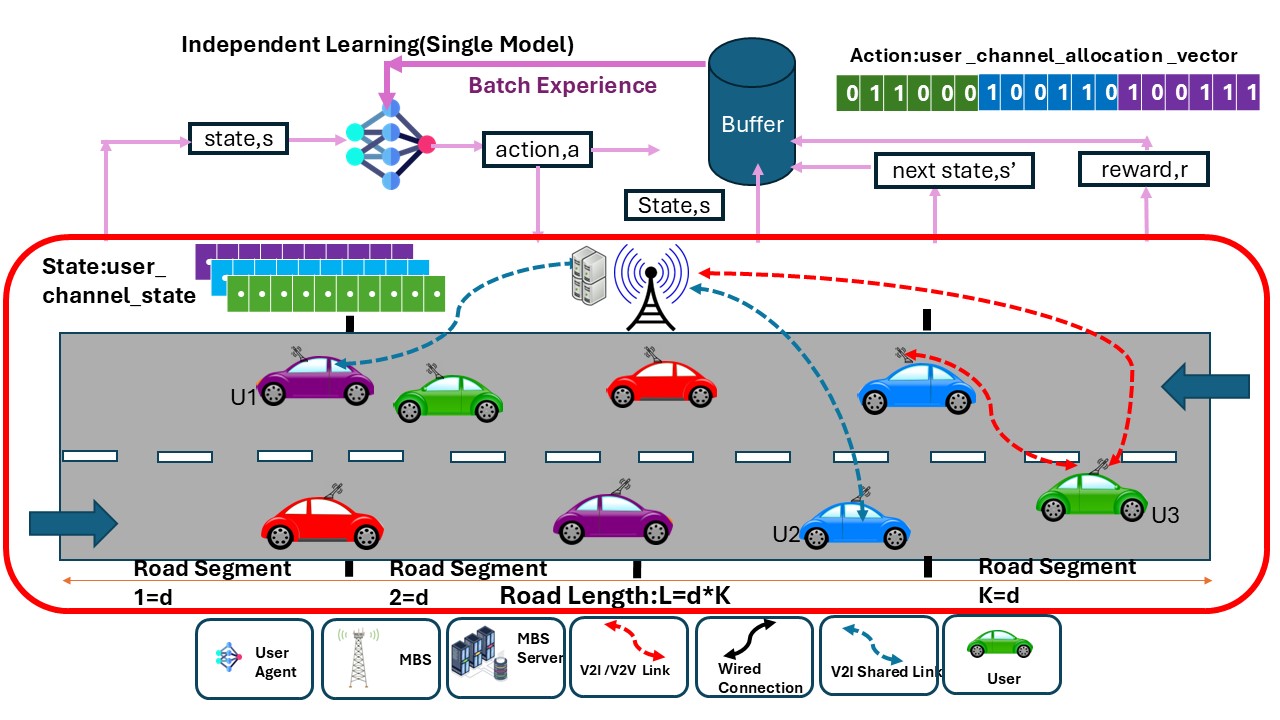}
    \caption{Typical Central Offloading with single-agent system.}
    \label{fig:single_server_single_agent_system}
\end{figure}
\begin{figure}[!t]
    \centering
    \includegraphics[width=\linewidth]{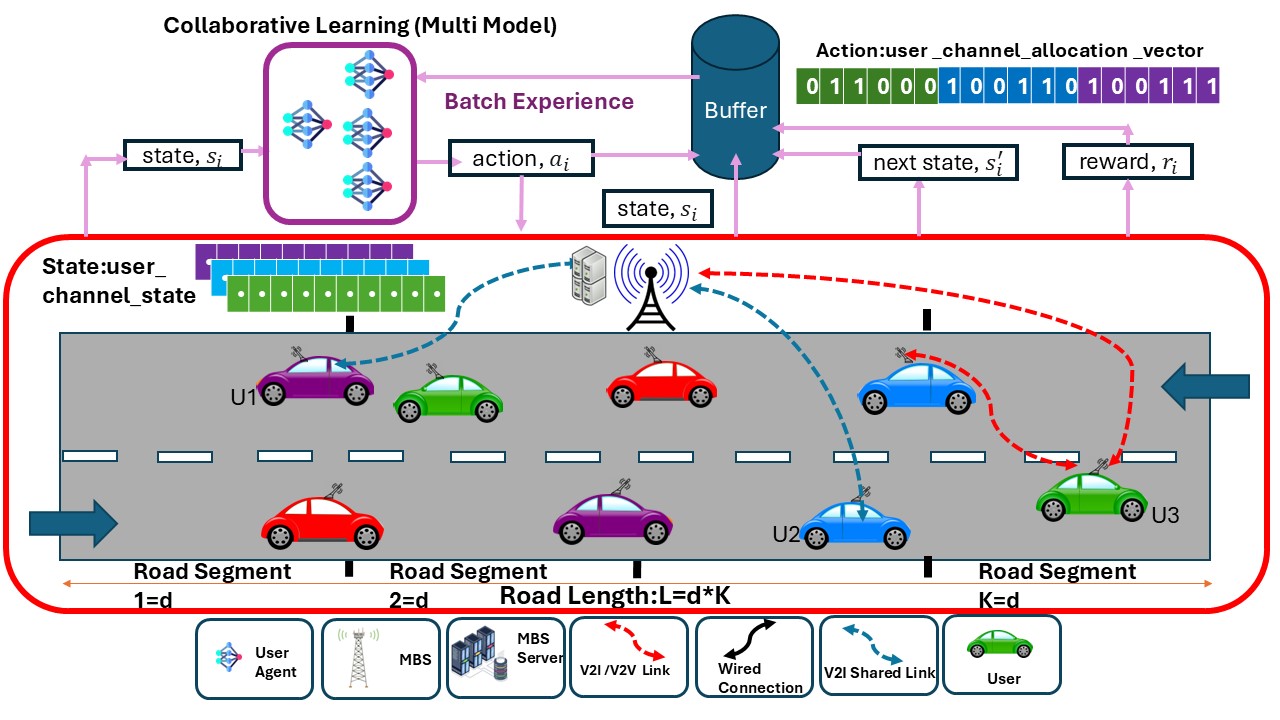}
    \caption{Typical Central Offloading with multi-agent system.}
    \vspace{-0.75em}
    \label{fig:single_server_multi_agent_system}
\end{figure}

\section{Centralized Offloading: Single-Server Computation Architectures}
\label{sec:single_server}
Having examined the fundamental deep reinforcement learning techniques and multi-agent frameworks relevant to vehicular edge computing, we now introduce how these methods have been applied in centralized offloading models, where a single MEC server centrally optimizes offloading decisions and overall system performance.
\subsection{System Overview}
Centralized offloading refers to a system architecture in which a single central server, typically located at an MBS, handles all computational task offloading requests from multiple user devices. In this model, the central entity is responsible for both task execution and resource allocation. User devices such as smartphones, IoT nodes, or vehicles typically offload their computational tasks to the central server due to limited onboard processing capabilities and energy constraints.

Fig.~\ref{fig:single_server_single_agent_system} illustrates a centralized DRL offloading framework employing a single-agent architecture. A central DRL agent, co-located with the MBS, makes offloading decisions on behalf of multiple user vehicles. Each vehicle observes local environmental parameters, such as wireless channel conditions and task attributes, and forwards this information to the central agent, which encodes it into a unified global state vector. Based on this input, the agent generates an action vector from a learned policy, typically a binary representation indicating spectrum band allocations for each user. The design of the reward function in this framework depends on whether spectrum reuse is allowed. If frequency reuse is disallowed, the reward penalizes assigning the same spectrum band to multiple users. In contrast, when reuse is supported, an interference-aware model must be incorporated to calculate rewards based on metrics such as data rate, latency, and energy consumption. As the environment transitions between states in response to actions taken, the central agent iteratively refines its policy to improve global system performance.

Fig.~\ref{fig:single_server_multi_agent_system} depicts a decentralized multi-agent DRL variant within the same centralized offloading setting. Here, each vehicle independently acts as a learning agent, observing its own local state (e.g., channel gain, battery level) and using a locally trained policy to select its desired spectrum allocation. These local action requests are sent to the MBS over V2I or V2V links. The MBS then aggregates the states and actions of all agents and resolves conflicts by learning the joint spectrum-sharing dynamics. The reward signal can be structured either as a global system objective or as a composition of individual agent rewards, depending on whether coordination or fairness is prioritized. Each agent subsequently updates its own policy based on the received feedback, promoting cooperative spectrum sharing and distributed learning.

This family of architectures, studied in works such as~\cite{b3,b27,b32,b33,b35,b36,b42,b45}, offers a simplified system design by centralizing control over offloading decisions and resource management, while minimizing the need for complex peer-to-peer negotiation among devices.

\subsection{Key Offloading Techniques}

A variety of techniques have been proposed to model and optimize offloading decisions in centralized architectures. These approaches address distinct challenges such as energy efficiency, channel contention, and task heterogeneity. The following are key strategies explored in the literature for offloading to a single central server:

\begin{itemize}[wide=0pt]
  \item \textbf{Contract Theory:}  
    Vehicles negotiate resource-allocation contracts with the edge server to balance energy consumption against computational load, enabling incentive-aligned decision-making~\cite{b3}.

  \item \textbf{Spectrum Sharing and Sensing:}  
    Joint V2V/V2I spectrum allocation schemes improve bandwidth utilization by coordinating offloading and communication~\cite{b27}. Additionally, cooperative spectrum sensing is employed to identify idle channels before initiating offloading~\cite{b36}.

  \item \textbf{Adaptive Mode Selection:}  
    Offloading policies dynamically switch between local execution, edge computing, and cloud-based processing depending on real-time network states, such as congestion or delay thresholds~\cite{b32}.

  \item \textbf{Probabilistic Action Generation:}  
    Stochastic action sampling mechanisms are integrated into DRL models (e.g., DQN) to handle uncertainty and promote exploration in environments with fluctuating channel or task conditions~\cite{b33}.

  \item \textbf{Hierarchical Decomposition:}  
    Complex offloading problems, such as joint beamforming and task assignment, are decomposed into tractable convex subproblems and coordinated using an outer DRL framework~\cite{b35}.

  \item \textbf{QoS-Driven Resource Estimation:}  
    Flow-level estimators are employed to map application-specific QoS requirements to resource allocation decisions, ensuring latency, throughput, or reliability targets are met~\cite{b42}.

  \item \textbf{Integrated Game Theory with MARL:}  
    Hybrid models combine competitive game-theoretic formulations with cooperative MARL frameworks to manage contention and resource conflicts among multiple users~\cite{b45}.

  \item \textbf{DAG-Based Task Scheduling:}  
    Task dependencies are modeled using DAGs, enabling structured scheduling and offloading of subtasks, particularly in UAV-based edge computing setups~\cite{b76}.
\end{itemize}

\subsection{DRL Paradigms}
The surveyed offloading frameworks can be broadly categorized into \emph{single-agent} and \emph{multi-agent} DRL paradigms, depending on whether policy learning is conducted by an individual learner or a set of distributed agents. In the single-agent setting, each user device or vehicle independently interacts with the environment and learns an offloading policy in isolation, assuming a fixed central edge server. These approaches often adopt value-based methods: DQN and their variants are widely used for handling discrete offloading actions, as demonstrated in~\cite{b3,b36,b42}. For continuous control tasks, such as adjusting transmission power or computation frequency, policy-based methods like PPO have been applied~\cite{b76}. Actor-critic frameworks such as DDPG further enable learning over high-dimensional continuous action spaces, allowing fine-grained control over offloading parameters~\cite{b35,b52}.

In contrast, multi-agent DRL methods assume that multiple devices, such as vehicles, UAVs, or IoT nodes, simultaneously learn and coordinate their offloading strategies, even when targeting a common central server. Under the CTDE paradigm, actor-critic algorithms such as MADDPG and MAPPO allow agents to share global information during training and execute independently at run time, facilitating cooperative behavior~\cite{b45,b76}. Alternatively, value-based game-theoretic approaches extend multi-agent DQN by integrating competitive or cooperative reward structures derived from game theory~\cite{b27,b78}. These multi-agent frameworks are particularly effective in capturing inter-device dynamics, such as resource contention and load balancing, yielding more robust and adaptive offloading strategies in dense or highly interactive environments.

\subsection{Optimization Objectives}
Across both single-agent and multi-agent DRL formulations, the central goal of offloading optimization is to improve system performance under realistic constraints. The most common objective is \emph{energy minimization}, where DRL agents learn policies that reduce energy consumption at both the device and network levels~\cite{b3,b32,b35}. This is particularly important for battery-constrained user equipment in vehicular and mobile environments.

For latency-sensitive applications, such as real-time sensing, augmented reality, or vehicular safety, \emph{end-to-end delay minimization} becomes the primary objective. In these cases, DRL agents are trained to make offloading decisions that reduce cumulative transmission and computation latency~\cite{b27,b76}. Additional objectives include \emph{resource utilization} and \emph{fairness}, where the goal is to balance the computational and communication loads across users and edge servers. This has been addressed using flow-level QoS estimators and reward shaping aligned with service-level agreements~\cite{b42}. In security-sensitive settings, \emph{resilience and reliability} are also critical: DRL agents are trained to maintain robust offloading decisions under adversarial conditions such as eavesdropping, jamming, or server compromise~\cite{b27}. By embedding these diverse optimization goals into their reward functions, DRL-based methods demonstrate high flexibility in adapting to heterogeneous application requirements and deployment constraints.

\begin{table*}[!t]
  \centering

  \caption{Comparison of DRL-Based Centralized Offloading Schemes}
  \label{tab:single_comparison}
    \begin{tabular}{|c|c|>{\centering}m{1.25cm}|m{3cm}|m{5cm}|c|}
      \hline
      \makecell[c]{\textbf{Agent}\\\textbf{Type}} &
      \textbf{Ref.} &
      \makecell[c]{\textbf{DRL}\\\textbf{Method}} &
      \makecell[c]{\textbf{Optimization}\\\textbf{Objective}} &
      \makecell[c]{\textbf{Key}\\\textbf{Techniques}} &
      \makecell[c]{\textbf{Computing}\\\textbf{Source}} \\
      \hline
      \hline

      \multirow{6}{*}{Single Agent}
        & \cite{b3}  & \multirow{4}{*}{DQN} &
          Energy, Latency &
          Contract Theory for Offloading Incentives &
          Vehicle, Fog \\
      \cline{2-2}\cline{4-6}
        & \cite{b32} & ~ &
          Energy &
          Adaptive Mode Selection for MTC &
          Edge, Cloud \\
      \cline{2-2}\cline{4-6}
        & \cite{b33} & ~ &
          Energy, Data Rate &
          Probabilistic Action Sampling &
          Local, Edge \\
      \cline{2-2}\cline{4-6}
        & \cite{b42} & ~ &
          Delay, Resource Utilization &
          QoS-Driven Resource Allocation &
          Edge \\
      \cline{2-6}
        & \cite{b35} & DDPG &
          Energy Efficiency &
          Hierarchical Learning &
          Local, Edge \\
      \cline{2-6}
        & \cite{b52} & TD3 &
          Computation Rate &
          RIS Phase Shift Control and Energy Partitioning &
          Local, Edge (BS) \\
      \hline

      \multirow{5}{*}{Multi Agent}
        & \cite{b45} & \makecell{MADDPG \\ (CTDE)} &
          Delay &
          Game-Theoretic Multi-Agent Coordination &
          Local, Edge \\
      \cline{2-6}
        & \cite{b78} & \makecell{MAAC \\ (DTDE)} &
          Latency, Computation Load &
          DNC Mechanism &
          Edge \\
      \cline{2-6}
        & \cite{b27} & \makecell{MADDQN \\ (DTDE)} &
          Latency &
          Joint V2V/V2I Spectrum Sharing &
          Edge (BS) \\
      \cline{2-6}
        & \cite{b76} & \makecell{MATD3 \\ (DTDE)} &
          Latency Reduction &
          DAG-Based UAV Task Scheduling &
          UAV, Local \\
      \cline{2-6}
        & \cite{b36} & \makecell{MADQN \\ (DTDE)} &
          Energy, Queue Length &
          Cooperative Spectrum Sensing &
          Local, Edge \\
      \hline
    \end{tabular}%
    \vspace{-0.75em}
\end{table*}

Table~\ref{tab:single_comparison} summarizes representative DRL-based centralized offloading schemes, comparing their agent configurations, optimization objectives, employed DRL algorithms, core techniques, and computing sources.

\subsection{Challenges and Limitations}
Despite notable advancements, existing DRL-based offloading architectures exhibit several recurring limitations. These challenges span modeling fidelity, evaluation rigor, and real-world scalability. This subsection categorizes these bottlenecks and outlines potential directions for future improvement.
\subsubsection{Modeling and Representation}
Several studies suffer from incomplete or overly abstract state and reward formulations. For instance, \cite{b32} omits the energy cost associated with data transmission from the terminal device to the MEC server in the reward function, leading to an underestimation of overall energy consumption. Similarly, \cite{b33} fails to incorporate task and resource dynamics into the state representation, reducing adaptability to real-time system variations. \cite{b3} oversimplifies user-specific and location-specific attributes, neglecting dynamic environmental factors such as mobility and load distribution, which can substantially affect offloading performance. Likewise, \cite{b45} overlooks the modeling of channel and server dynamics, restricting the agent's ability to account for temporal fluctuations in network conditions. Accurate modeling of such dynamics is critical to improving generalization and policy robustness in highly variable vehicular networks.

\subsubsection{Evaluation and Benchmarking}
Many existing works exhibit limitations in their experimental validation. \cite{b36} reports uniform convergence during training, raising concerns about deterministic behavior in phases where stochastic exploration is expected. In \cite{b76}, agents demonstrate varying convergence speeds, indicating the need for asynchronous learning rates and temporal scaling when coordinating heterogeneous agents. \cite{b27} adopts fixed network parameters (e.g., static exploration rates), which limit adaptability to dynamic environments. Additionally, \cite{b3} lacks comparative evaluation against baseline DRL algorithms, which hinders objective assessment of performance gains. That work also omits QoS-centric metrics such as latency and throughput, instead relying solely on utility-based visualization. Without thorough benchmarking, it becomes difficult to assess real-world applicability or trade-offs between performance metrics.

\subsubsection{Scalability and Deployment}
Although DRL-based offloading systems perform well in simulation or controlled environments, their deployment at scale remains challenging. High-dimensional action spaces pose significant computational burdens, especially in multi-agent or resource-constrained settings. Furthermore, many models lack the flexibility to accommodate dynamic user participation or topology changes. Achieving real-time inference on edge devices also demands lightweight, low-latency models. Future research must focus on bridging the gap between theoretical models and practical deployment by designing scalable, efficient, and adaptive DRL frameworks that remain robust in decentralized, multi-agent vehicular edge networks.

\subsection{Conclusion}
Centralized offloading remains a compelling strategy for managing computational tasks in environments with limited local processing resources. A wide array of DRL-enabled techniques, including contract theory, spectrum sharing, hierarchical optimization, and QoS-driven resource allocation, have shown promise in enhancing energy efficiency, reducing latency, and improving overall system performance. However, several challenges remain. Incomplete MDP representations, insufficient baseline comparisons, and inadequate exploration mechanisms may reduce the robustness and generalizability of proposed methods. To achieve reliable and scalable deployment, future work must address these limitations by incorporating more realistic system dynamics, adopting rigorous evaluation frameworks, and developing lightweight, adaptive learning models that can operate under the complexities of real-world vehicular edge environments.

\begin{figure}[!t]
  \centering
    \includegraphics[width=\linewidth]{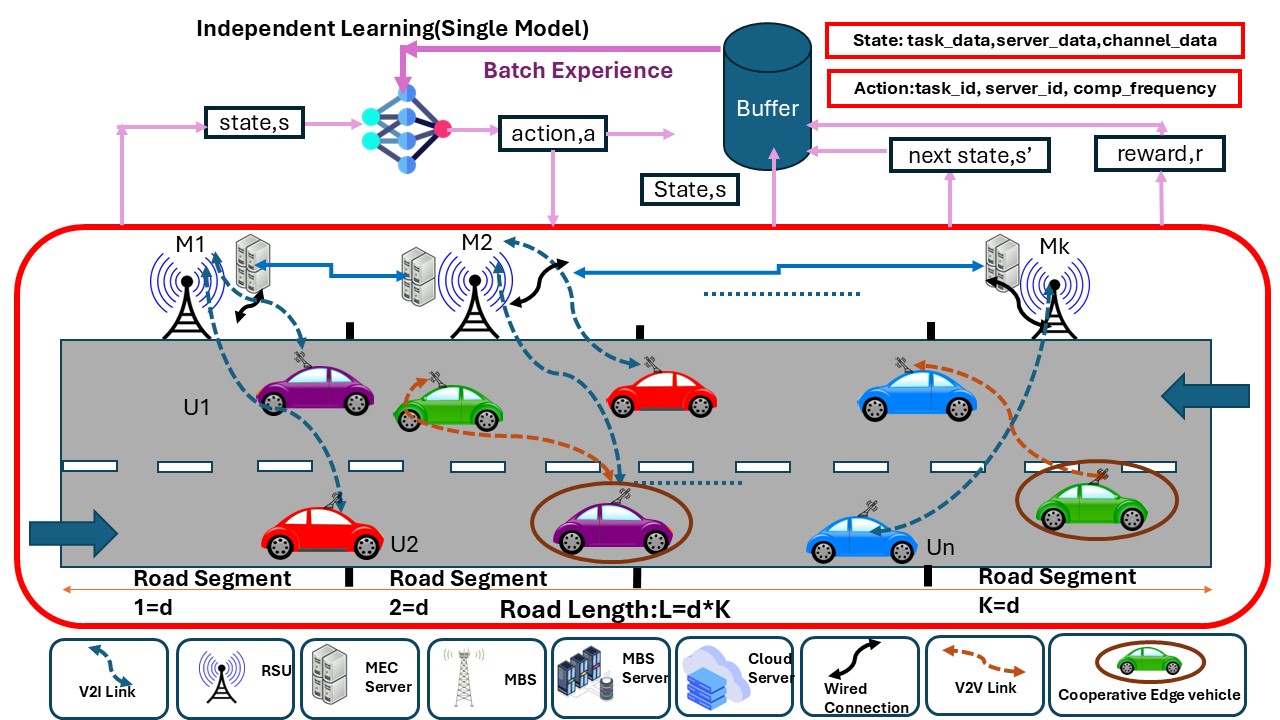}
    \caption{Typical Distributed Offloading with single-agent system.}
    \label{fig:multi_server_single_agent_system}
\end{figure}
\begin{figure}[!t]
    \centering
    \includegraphics[width=\linewidth]{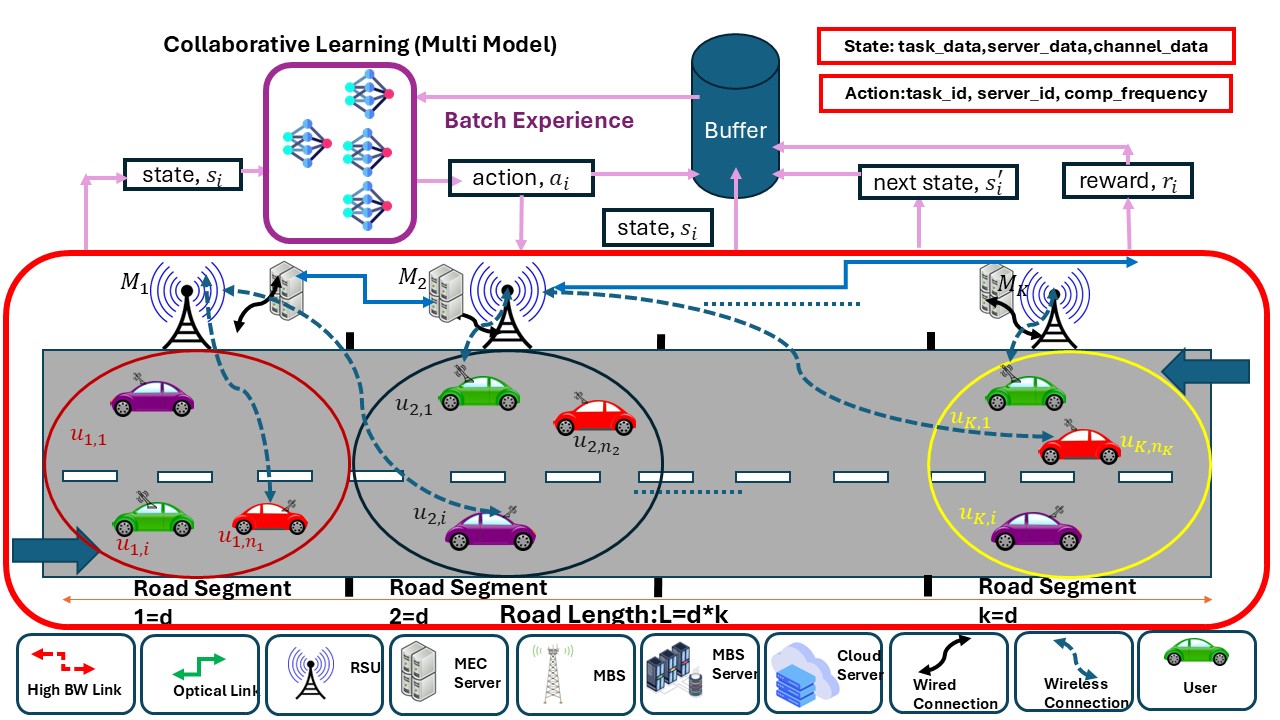}
    \caption{Typical Distributed Offloading with multi-agent system.}
    \vspace{-0.75em}
    \label{fig:multi_server_multi_agent_system}
\end{figure}

\section{Distributed Offloading: Multi-Server Cooperative Architectures}
\label{sec:multi_server}
While centralized offloading architectures enable global coordination, their reliance on a single control or resource entity introduces scalability, processing burden, and signaling limitations in highly dynamic vehicular environments. Consequently, distributed offloading approaches have emerged as a promising alternative, enabling decentralized decision making based on local observations.
\subsection{System Overview}
Unlike the centralized offloading paradigm, where a single server manages all computational tasks, distributed offloading leverages a network of interconnected servers or edge nodes that collaboratively handle the computational workload. In this architecture, servers exchange information about network conditions, resource availability, and task requirements to make joint decisions that optimize system-wide performance. This decentralized cooperation enhances scalability, fault tolerance, and responsiveness in complex vehicular environments.

Fig.~\ref{fig:multi_server_single_agent_system} illustrates a single-agent distributed offloading architecture, where one central RL agent manages decision-making across all $K$ servers (e.g., associated with $K$ road segments). The environment is modeled as a unified MDP, and the agent determines optimal actions for computational task offloading based on aggregated observations such as channel quality, task load, and available server resources. At each time step, the agent selects a subset of users, typically one per road segment, for computational task offloading. Actions may include server selection, offloading decision, and computational frequency configuration. State transitions, actions, and rewards are stored in a shared replay buffer to enable centralized learning. While this setup simplifies system modeling and training, it may fail to capture the full complexity of distributed dynamics and can deviate from globally optimal performance in highly variable settings.

In contrast, the multi-agent distributed offloading architecture shown in Fig.~\ref{fig:multi_server_multi_agent_system} models each of the $K$ servers as an independent learning agent. This formulation gives rise to $K$ distinct POMDPs, where each agent observes only local environmental states within its respective service zone. Agents select actions autonomously based on their local conditions, including user demands, channel state, and server load. During training, however, agents share experience via a centralized buffer, following the CTDE paradigm. This collaborative framework enables agents to learn inter-agent dynamics, minimize conflict, and jointly optimize resource allocation. Although more computationally intensive, this multi-agent setup provides superior adaptability and scalability in large-scale, spatially distributed, and time-varying environments.

This class of distributed offloading architectures has been explored in numerous studies, including~\cite{b5,b6,b7,b11,b82,b4,b10,b12,b15,b16,b19,b20,b23,b24,b29,b31,b41,b43,b53,b65,b62,b73,b74,b64,b61,b68,b71,b72,b63,b66}.

\subsection{Key Offloading Techniques}
A variety of complementary techniques have been proposed to structure distributed offloading problems beyond core DRL formulations:

\begin{itemize}[wide=0pt]
  \item \textbf{Cooperative Vehicle Selection:}  
  \cite{b5} employs historical vehicle trajectories and LightGBM to select edge nodes for computational task offloading. \cite{b6} extends this by using XGBoost for enhanced selection robustness. \cite{b7} incorporates link-duration estimation to improve the reliability of offloading connections.

  \item \textbf{Task-Adaptive Framing and Prioritization:}  
  \cite{b11} introduces a dynamic framing method that adjusts subtasks based on real-time system conditions. \cite{b82} implements a prioritization strategy to enable adaptive task selection for offloading.

  \item \textbf{Task Migration and Handover:}  
  \cite{b15,b65} focus on dynamic rescheduling mechanisms between vehicles to support task migration. \cite{b23} addresses seamless handover within a distributed edge system. \cite{b31} integrates task partitioning with DRL, considering task types and vehicle speed in the decision process.

  \item \textbf{Attention and Hierarchical Models:}  
  \cite{b62} uses an attention-based BiLSTM model to capture temporal dependencies in LEO network environments. \cite{b73} applies a game-theoretic hierarchical DRL approach for UAV-based service assignment. \cite{b74} proposes action-branching network structures to address challenges posed by high-dimensional action spaces.

  \item \textbf{Social and Mobility Awareness:}  
  \cite{b16} leverages social interactions to support content caching and offloading strategies. \cite{b19} adapts offloading policies based on observed vehicle mobility patterns.

  \item \textbf{Topology and Clustering Innovations:}  
  \cite{b61} organizes vehicles into fog-based clusters to facilitate V2V and V2R offloading. \cite{b68} optimizes task partitioning in ultra-dense network settings. \cite{b71} integrates HAP-supported communication for enhanced power efficiency, while \cite{b72} applies knowledge-based multi-agent methods in UAV networks.
\end{itemize}

\subsection{DRL Paradigms}
The surveyed distributed offloading systems apply various DRL paradigms, broadly categorized into single-agent and multi-agent approaches.

In single-agent frameworks, each device independently learns its offloading policy. Value-based methods such as DQN and adaptive priority DQN are used for discrete action spaces~\cite{b4,b5,b82}, while variants like Double DQN and Dueling DQN are employed to improve training stability and mitigate Q-value overestimation~\cite{b6,b10,b11,b12}. For continuous action spaces, SAC has been used effectively~\cite{b7}.

In multi-agent frameworks, several learners interact and coordinate their decisions during training or execution. Actor-critic methods such as MADDPG and MAPPO follow the CTDE paradigm~\cite{b29,b31,b68,b73}. Graph-based MADDPG is used in~\cite{b24}, while MAAC with cost-revenue modeling appears in~\cite{b41}. COMA is applied for counterfactual credit assignment in cooperative tasks~\cite{b61,b62,b65}. Value decomposition techniques such as VDN and QMIX support scalable learning by separating the joint Q-function into local components~\cite{b63,b64,b74}. Fully decentralized frameworks include hybrid approaches using DDPG and SAC~\cite{b71,b72}, as well as decentralized variants of MADQN and MAAC that target specific applications such as task migration and joint latency-compute optimization~\cite{b23,b78}.

\subsection{Optimization Objectives}
The distributed DRL-based offloading schemes target a range of optimization goals, depending on application requirements and system constraints. One of the most critical objectives is \emph{latency minimization}, particularly for real-time applications. Various approaches aim to reduce task completion delay through mechanisms such as digital-twin scheduling~\cite{b4}, dynamic subtask framing~\cite{b11}, and handover-aware migration strategies~\cite{b23,b65}. Additional methods consider mobility-aware DRL~\cite{b19} and UAV trajectory planning~\cite{b73}, which adapt decisions based on vehicular movement and aerial link quality. Quality-of-service (QoS) driven RSU selection is also employed to maintain timely service delivery in dynamic roadside environments~\cite{b61}.

A second major focus is \emph{joint energy-latency optimization}, where reward functions balance power consumption with delay requirements. Examples include ID-pool balancing~\cite{b10}, hybrid frameworks combining DDPG and SAC~\cite{b71}, and federated MADDPG with interference coordination~\cite{b29}, each demonstrating effectiveness in managing both energy use and latency. Additional strategies such as priority-weighted DQN and delay-constrained energy management~\cite{b82,b62} adjust exploration behavior and cost penalties to meet specific timing and energy budgets. Several works also address \emph{QoS and QoE improvement}, integrating models based on cost-revenue trade-offs and user significance~\cite{b41}, heterogeneous offloading strategies for enhanced user experience~\cite{b43}, and fairness-aware resource allocation using data-rate optimization and delay constraints~\cite{b63,b64,b72}. These objectives demonstrate the flexibility of DRL in tailoring offloading decisions to support diverse operational goals and service requirements.

\begin{table*}[!t]
  \centering

  \caption{Comparison of Distributed DRL-Based Offloading Schemes}
  \label{tab:distributed_comparison}
    \begin{tabular}{|c|c|>{\centering}m{1.4cm}|>{\centering}m{2.5cm}|c|c|}
      \hline
      \makecell[c]{\textbf{Agent}\\\textbf{Type}} &
      \textbf{Ref.} &
      \makecell[c]{\textbf{DRL}\\\textbf{Method}} &
      \makecell[c]{\textbf{Optimization}\\\textbf{Objective}} &
      \makecell[c]{\textbf{Key}\\\textbf{Techniques}} &
      \makecell[c]{\textbf{Computing}\\\textbf{Source}} \\
      \hline\hline

      \multirow{14}{*}{Single Agent}
        & \cite{b4}  & \multirow{3}{*}{DQN}     & Delay                                & Digital Twin Scheduling                              & Local, Edge \\ \cline{2-2}\cline{4-6}
        & \cite{b5}  & ~     & Data Rate, Resource Utilization       & LightGBM-based Cooperative Vehicle Selection          & Vehicular \\ \cline{2-2}\cline{4-6}
        & \cite{b82} & ~     & Latency, Energy, Completion Ratio     & Priority-Based Task Selection                         & Edge \\ \cline{2-6}
        & \cite{b10} & \multirow{3}{*}{DDQN}    & Latency, Energy, Cost                 & Shared Result Pooling with ID Grouping                & Local, Edge, BS \\ \cline{2-2}\cline{4-6}
        & \cite{b11} & ~    & Delay                                & Subtask Framing for Dynamic Offloading                & Local, Edge, BS \\ \cline{2-2}\cline{4-6}
        & \cite{b12} & ~    & Energy, Latency                      & Handover-Aware Partial Offloading                     & Local, Vehicular, Edge \\ \cline{2-6}
        & \cite{b6}  & \multirow{2}{*}{D3QN}    & Latency, Resource Utilization        & Cooperative Selection with XGBoost                    & Vehicular \\ \cline{2-2}\cline{4-6}
        & \cite{b43} & ~    & QoE                                  & Task-Type-Aware Heterogeneous Offloading              & Local, Edge \\ \cline{2-6}
        & \cite{b15} & \multirow{4}{*}{DDPG}    & Latency                              & Partitioned Offloading, Migration with DRL         & Edge \\ \cline{2-2}\cline{4-6}
        & \cite{b16} & ~    & Latency                              & Social-Aware Content Caching                          & Vehicle, Edge \\ \cline{2-2}\cline{4-6}
        & \cite{b19} & ~    & Energy, Latency                      & Mobility-Aware Task Scheduling                        & RSU, Local \\ \cline{2-2}\cline{4-6}
        & \cite{b20} & ~    & Energy, Latency                      & Trust Evaluation using CNN, GNN, LSTM                 & Local, Edge \\ \cline{2-6}
        & \cite{b53} & TD3     & Energy, Latency                      & Cooperative RSU Migration, Load Transfer           & Local, Edge \\ \cline{2-6}
        & \cite{b7}  & SAC     & Latency, Cost                        & Cooperative Vehicle Selection                         & Vehicle, Edge \\
      \hline

      \multirow{13}{*}{Multi Agent}
        & \cite{b23} & MADQN (DTDE)   & Delay-Constrained Migration Cost     & Handover-Aware Task Migration                         & Local, Edge \\ \cline{2-6}
        & \cite{b29} & \multirow{4}{*}{\makecell[c]{MADDPG\\ (CTDE)}} & Energy, Latency                     & Federated Learning, Interference Coordination      & Local, Edge \\ \cline{2-2}\cline{4-6}
        & \cite{b31} & ~ & Energy, Latency                     & Task-Type and Vehicle-Speed-Aware Scheduling          & Local, Vehicular, Edge \\ \cline{2-2}\cline{4-6}
        & \cite{b68} & ~ & Latency, Computation Bit, Energy   & Joint Task Partitioning and Resource Allocation       & Local, Edge \\ \cline{2-2}\cline{4-6}
        & \cite{b24} & ~ & Latency-Constrained Cost           & Graph-Based Multi-Agent Coordination                  & Vehicular \\ \cline{2-6}
        & \cite{b73} & Game MADDPG   & Latency, Fairness                  & Game-Theoretic Hierarchical Offloading                & UAV, Local \\ \cline{2-6}
        & \cite{b41} & MAAC (CTDE)     & QoS                                & Cost-Revenue Modeling, User Significance           & Local, Edge \\ \cline{2-6}
        & \cite{b71} & MADDPG + SAC   & Latency, Energy                    & Integrated MADDPG with SAC                            & Local, HAP \\ \cline{2-6}
        & \cite{b72} & MASAC (CTDE)    & Latency, Fairness                  & Knowledge-Guided Policy Exploration                   & Local, UAV \\ \cline{2-6}
        & \cite{b66} & MAA2C (CTDE)    & Latency                            & MDP Approximation of Non-Cooperative POMDP            & Local, Edge \\ \cline{2-6}
        & \cite{b61} & \multirow{3}{*}{\makecell[c]{COMA\\ (CTDE)}}   & Latency                            & Cross-Region Offloading via Service Vehicle Clusters  & Service Vehicle \\ \cline{2-2}\cline{4-6}
        & \cite{b65} & ~   & Latency                            & Task Migration Strategy                               & Edge \\ \cline{2-2}\cline{4-6}
        & \cite{b62} & ~   & Energy                              & BiLSTM-Based Attention for LEO Systems                & Satellite \\
      \hline
    \end{tabular}%
  \vspace{-0.75em}
\end{table*}

Table~\ref{tab:distributed_comparison} provides a comparative overview of distributed DRL-based offloading schemes, highlighting agent configurations, optimization objectives, core algorithmic strategies, and associated computing infrastructures.

\subsection{Challenges and Limitations}
Despite notable progress, DRL-based distributed offloading frameworks continue to face several critical challenges. This subsection categorizes the most common limitations and outlines directions for improvement.

\subsubsection{Modeling and Representation}
Many studies adopt simplified MDP models that fail to capture key dynamics of vehicular edge environments. For instance, \cite{b32} focuses on energy consumption but overlooks the transmission energy cost between terminal devices and MEC servers. \cite{b45} neglects channel fluctuations and user mobility in its state space representation. Similarly, \cite{b19} ignores task urgency by treating all offloading requests equally, while \cite{b63} uses rigid penalty terms (e.g., for SNR or collisions), which may dominate the reward function and hinder balanced learning. These oversights reduce the robustness and adaptability of learned policies. Improved modeling should incorporate variables such as transmission energy, channel dynamics, mobility patterns, and more flexible penalty structures to reflect real-world variability.

\subsubsection{Evaluation and Benchmarking}
Several works exhibit shortcomings in performance evaluation and baseline comparisons. For example, \cite{b5} compares tabular SARSA to deep RL without using a comparable deep baseline. \cite{b11} reports that DDQN outperforms Dueling DQN but provides no explanation for the result. Some studies, such as \cite{b27} omit head-to-head comparisons with standard DRL methods, while others report only utility values without examining critical metrics like latency or energy consumption~\cite{b7}. Instability in reward trajectories has been observed~\cite{b10}, and varying results under different priority settings are reported in~\cite{b29,b31}. Overreliance on immediate rewards can result in shortsighted policies~\cite{b19}, and the use of expert-designed features limits generalizability~\cite{b72}. Uniform communication models, as in \cite{b73}, mask complex action-state dependencies, and sparse exploration in value-decomposition methods~\cite{b74} may reduce their effectiveness. Future evaluations should include consistent DRL baselines, full reporting of QoS and QoE metrics, and in-depth convergence analyses across diverse operating conditions.

\subsubsection{Deterministic vs. Stochastic Policies}
Deterministic policies, such as those used in DDPG and MAPPO, are often applied to continuous control problems. However, their use in discrete or hybrid action spaces introduces unresolved challenges. For instance, \cite{b15}, \cite{b29}, and \cite{b31} do not detail how continuous outputs are discretized or how exploration is maintained. Although noise injection is a common method for encouraging exploration, its design (e.g., noise scale and decay rate) is often ad hoc and rarely evaluated systematically. A more principled approach to discretization and noise design, potentially using hybrid stochastic-deterministic models, could improve training stability across varied decision spaces.

\subsubsection{Synchronization and Coordination}
Achieving coordinated behavior in multi-agent offloading remains a key difficulty. \cite{b12} reports issues in synchronizing partial computational task offloading, leading to inefficient resource use. \cite{b43} highlights contention when multiple users select the same edge server, degrading performance under congestion. \cite{b20} includes all features in an LSTM without prioritizing relevant inputs, reducing training efficiency, while \cite{b64} shows that information sharing alone does not ensure agent cooperation. Simplified POMDP-to-MDP transformations may distort actual transition dynamics~\cite{b66}. Even CTDE-based methods show limitations: \cite{b68} and \cite{b71} report difficulty reconciling global training with local execution, which leads to unstable convergence. Addressing these problems requires improved feature selection, adaptive synchronization, and hybrid training strategies that better align centralized and decentralized components.

\subsection{Conclusion}
In summary, distributed offloading strategies in vehicular networks and edge computing offer significant advantages in reducing latency, improving resource utilization, and enhancing scalability. Through collaboration among interconnected servers and agents, often supported by multi-agent learning frameworks, these architectures reduce bottlenecks and alleviate single-point failures. Nonetheless, several challenges persist. Incomplete MDP formulations weaken the robustness of decision-making, particularly when omitting critical dynamics such as transmission energy, mobility, or communication quality. Reward design remains problematic, especially when penalization is miscalibrated or fails to reflect task priorities. Evaluation practices often lack consistent baselines or detailed QoS comparisons, which limits confidence in reported improvements. Additionally, multi-agent architectures continue to face issues with synchronization, fairness, and coordination, particularly when managing shared resources in dynamic, decentralized environments. Addressing these limitations will require more comprehensive modeling, principled evaluation methods, and scalable DRL frameworks capable of adapting to real-world vehicular edge conditions.

\begin{figure}[!t]
  \centering
    \includegraphics[width=\linewidth]{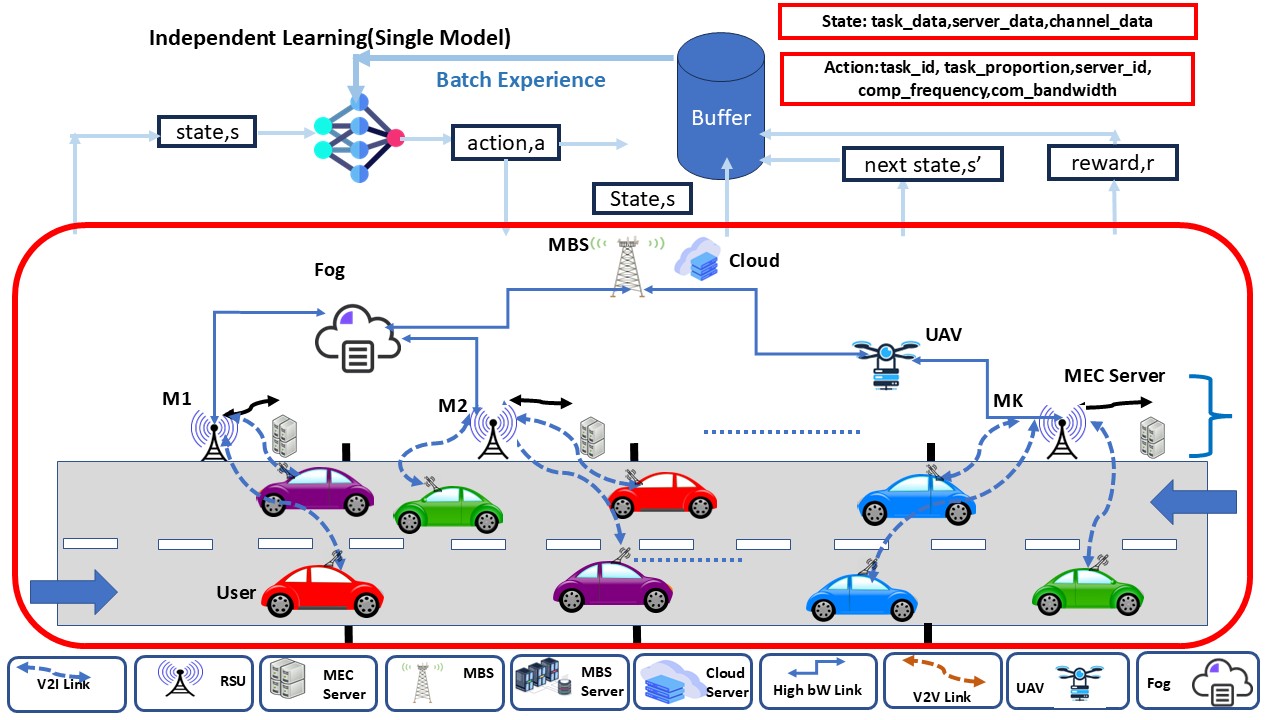}
    \caption{Typical Hierarchical Offloading with single-agent system.}
    \label{fig:het_single_agent_system}
\end{figure}
\begin{figure}[!t]
    \centering
    \includegraphics[width=\linewidth]{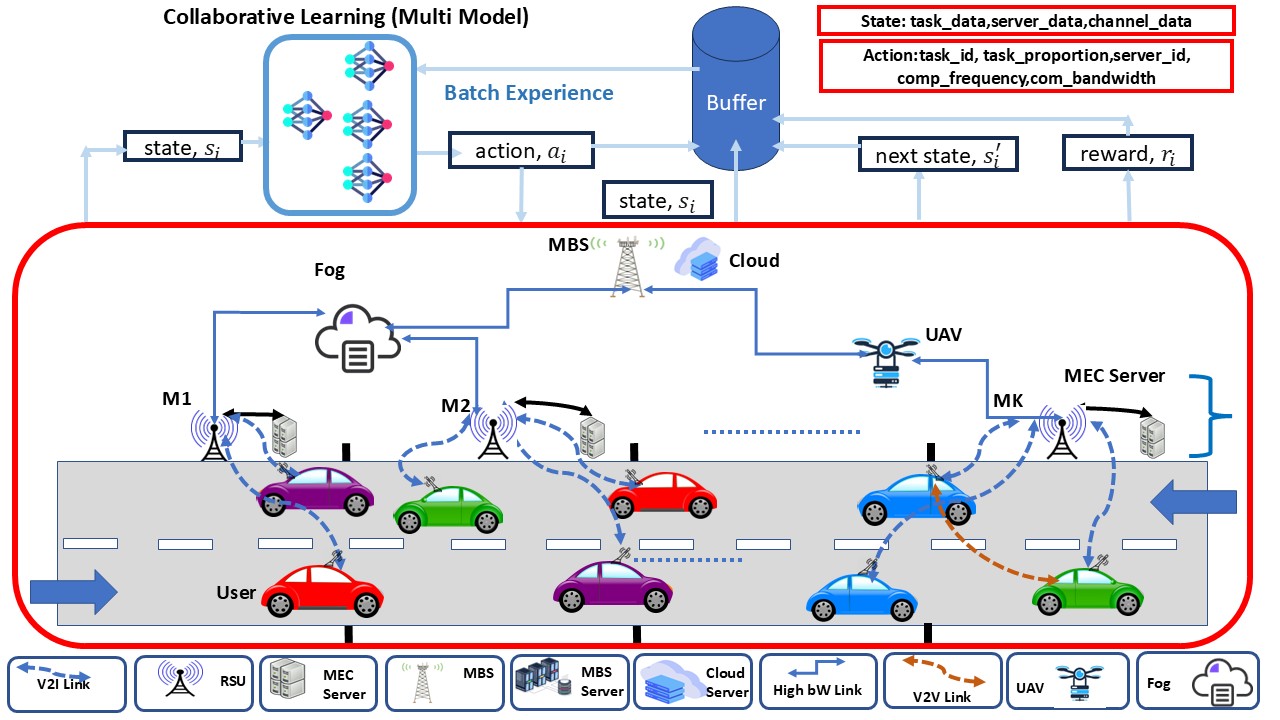}
    \caption{Typical Hierarchical Offloading with multi-agent system.}
    \vspace{-0.75em}
    \label{fig:het_multi_agent_system}
\end{figure}

\section{Hierarchical Offloading: Multi-Tier and Heterogeneous Ad Hoc Systems}
\label{sec:hierarchical}
Despite the scalability and adaptability achieved by distributed offloading architectures through decentralized decision making, limited global coordination can hinder system-level efficiency. Hierarchical or multi-tier offloading architectures address this limitation by combining local autonomy with higher-level coordination across edge, fog, UAV, and cloud layers.

\subsection{System Overview}
While distributed edge computing supports decentralized decision-making, it faces increasing coordination complexity, latency, and inefficiencies as network scale and task heterogeneity grow, particularly in dynamic vehicular environments. Independent decisions by individual nodes can lead to communication overhead, conflicting actions, and redundant computation. To address these limitations, hierarchical edge computing has emerged as a scalable alternative. It integrates heterogeneous computing resources deployed across multiple tiers, such as fixed MEC servers, fog nodes, cloud centers, and mobile platforms including UAVs, to enable more flexible and efficient task execution.

Fig.~\ref{fig:het_single_agent_system} illustrates a centralized offloading architecture employing a single-agent DRL framework. A central controller, typically located at an MBS, is responsible for managing task scheduling, resource allocation, and bandwidth assignment across all devices. A global state vector is constructed by aggregating task characteristics, channel conditions, and the availability of heterogeneous resources, including stationary and mobile edge servers. Based on this unified state, the central DRL agent generates an action vector specifying task ID, partition ratio, server selection, computational frequency, and bandwidth allocation. The agent receives performance feedback (e.g., latency, energy, or throughput) and updates its policy via experience replay. While this centralized model benefits from a comprehensive view of the system, it faces limitations in scalability and responsiveness under high dynamism.

Fig.~\ref{fig:het_multi_agent_system} presents a multi-agent DRL framework adapted to the same heterogeneous setting. Here, each vehicle functions as an autonomous agent with partial observability, forming its local state based on task attributes, local connectivity, and channel status. Each agent independently generates an action vector (including task partitioning, server selection, frequency configuration, and bandwidth request) and submits it to a central coordinator. Conflict resolution, such as spectrum contention or server overload, is handled centrally, while learning is conducted under the CTDE paradigm. This architecture improves scalability and responsiveness by distributing decision-making while retaining global coordination during training.

Both centralized and multi-agent approaches must contend with the complexity introduced by heterogeneous resource types. In UAV-enabled systems, trajectory optimization becomes a critical component of task assignment, while in ad hoc vehicular networks, cooperative peer selection requires the system to incorporate topology-aware and context-sensitive MDP formulations. These settings often involve multi-objective agents, each responsible for distinct, interdependent decisions. For example, a MEC server may simultaneously determine offloading, task migration, and frequency allocation, while a UAV must adapt its trajectory for coverage and reliability.

Such multi-tier architectures introduce high-dimensional, mixed-type action spaces, including discrete, continuous, and combinatorial decisions, making policy learning increasingly complex. In centralized systems, this results in scalability and convergence challenges. In contrast, decentralized systems allow each agent to specialize in a particular decision domain (e.g., UAVs handling mobility, MECs handling computation), which simplifies local policy learning. However, coordination remains essential to avoid conflicting or suboptimal global behaviors. This class of hierarchical and multi-tier DRL-based offloading systems has been explored in several studies, including \cite{b30,b34,b22,b2,b14,b38,b18,b25,b28,b67,b13,b44,b46,b69,b70,b75,b8,b17,b1,b21,b59,b51,b54,b55,b56,b57,b58}.

\subsection{Key Offloading Techniques}
Various strategies have been proposed to enhance task scheduling and resource allocation in hierarchical and heterogeneous offloading environments. The key techniques include:

\begin{itemize}[wide=0pt]
  \item \textbf{DDPG with Convex Optimization:}  
  \cite{b34} integrated DDPG with fully connected and convolutional networks, augmented by convex optimization solvers, to jointly optimize offloading decisions and energy consumption in complex system states.

  \item \textbf{CTDE:}  
  \cite{b30,b22} employed a CTDE framework, where a global critic facilitates policy learning, while decentralized agents execute local offloading actions across hierarchical infrastructures to achieve balanced load distribution.

  \item \textbf{Lightweight Independent IoT Agents:}  
  \cite{b38} deployed DRL-based agents directly on IoT devices, allowing real-time, decentralized offloading based on device-specific energy constraints and dynamic workload profiles.

  \item \textbf{Trust-Aware Vehicular Offloading:}  
  \cite{b2} introduced a trust-based strategy where social-awareness is embedded into the offloading decision process to identify and avoid unreliable peer nodes in vehicular networks.

  \item \textbf{Social-Relation-Based Clustering:}  
  \cite{b14} proposed clustering of vehicles based on social ties to enhance cooperative computational task offloading and resource reuse among proximate nodes.

  \item \textbf{Predictive Communication State Modeling:}  
  \cite{b25} utilized LSTM and bidirectional RNNs to model and predict link quality in dynamic environments, enabling anticipatory offloading based on future channel conditions.

  \item \textbf{Temporal Task Encoding via Seq2Seq BiGRU:}  
  \cite{b28} applied a sequence-to-sequence model with bidirectional GRUs for subtask allocation, capturing both temporal dependencies and task-specific features.

  \item \textbf{Collaborative Edge Offloading:}  
  \cite{b18} proposed a cooperative approach in which edge servers coordinate task allocation and resource sharing to minimize system latency.

  \item \textbf{Joint Offloading Policy Learning:}  
  \cite{b67} trained centralized policies to jointly determine offloading decisions between cloud and edge servers, optimizing task placement for performance and efficiency.

  \item \textbf{Fairness-Oriented Optimization:}  
  \cite{b46} incorporated fairness metrics into the reward structure to promote equitable task handling and balanced completion times across users.

  \item \textbf{DAG-Aware Multi-Agent Coordination:}  
  \cite{b44} represented task dependencies using directed acyclic graphs (DAGs) and enabled trajectory sharing among agents for improved collaborative scheduling.

  \item \textbf{Knowledge-Driven Policy Adaptation:}  
  \cite{b69,b70} leveraged historical task patterns and meta-learning to accelerate DRL policy adaptation in varying user contexts and application scenarios.

  \item \textbf{Adaptive Video Offloading:}  
  \cite{b13} developed a video quality control method that dynamically adjusts encoding parameters to reduce offloading cost while maintaining user experience.

  \item \textbf{Multi-Objective UAV-Based Scheduling:}  
  \cite{b75} designed a UAV-assisted offloading framework that simultaneously optimizes latency and energy consumption using a unified multi-objective reward function.
\end{itemize}

\subsection{DRL Paradigms}
In single-agent systems, a variety of DRL algorithms have been employed to enhance offloading and resource allocation efficiency. DQN and its variants are among the most commonly used. Distributed DQN~\cite{b8,b22} focuses on learning from distributed edge and UAV resources, whereas Double DQN~\cite{b2} integrates trust values into offloading decisions to enhance security. DDPG, when combined with fully connected neural networks and convex optimization~\cite{b34}, has demonstrated improvements in energy efficiency. Other task-specific applications include adaptive video quality control~\cite{b13} and socially-informed clustering for offloading~\cite{b14}. Cooperative edge processing~\cite{b18} is also supported by single-agent DRL to enhance system latency. To improve action exploration, \cite{b17} applies Ornstein-Uhlenbeck noise to DDPG, while Prime Dual DDPG~\cite{b21} optimizes task allocation between local and remote servers. A3C~\cite{b46} is used to minimize delays and promote fairness in scheduling. Distributed PPO~\cite{b44} leverages shared experience trajectories for improved coordination, and PPO~\cite{b28} enables smooth sequential task handling through policy refinement.

In multi-agent settings, Federated DQN (DTDE)~\cite{b1} allows agents to learn local models while sharing information to improve global optimization. Independent IoT agents~\cite{b38} allocate energy and manage tasks in a distributed manner, coordinating indirectly to resolve task conflicts. Double DQN (DTDE)~\cite{b2} incorporates trust-based mechanisms into latency- and energy-aware offloading policies. Joint action Q-learning~\cite{b67} enables agents to coordinate task decisions in shared environments. Fairness-driven strategies are addressed through SAC (DTDE)~\cite{b59}, which incorporates fairness into resource distribution decisions.

CTDE frameworks also play a key role. DDPG (CTDE)~\cite{b25} is applied alongside LSTM and bidirectional RNNs for communication state prediction and priority-aware task scheduling. MADDPG (CTDE)~\cite{b30,b70} focuses on latency and priority optimization using dynamic reward models. Additionally, decentralized A3C~\cite{b69} enables distributed learning using shared network weights across agents without requiring complete MDP visibility. Multi-agent actor-critic methods~\cite{b75} further advance decentralized DRL by supporting distributed decision-making in heterogeneous offloading environments.

\subsection{Optimization Objectives}
Various hierarchical offloading schemes have been proposed to meet critical performance goals such as minimizing latency, improving energy efficiency, and optimizing overall resource utilization. Several works have primarily targeted \textit{latency reduction}. For example, cooperative edge offloading~\cite{b18} and Prime Dual DDPG~\cite{b21} effectively reduced task processing delays by optimizing task distribution between edge and cloud resources. Additionally, latency minimization has been the focus in~\cite{b14,b28,b30,b44,b46,b67,b69}, where task scheduling and resource allocation strategies were developed to ensure timely task execution within multi-tier and heterogeneous infrastructures. Other approaches emphasize \textit{joint energy-latency optimization}. Studies such as~\cite{b1,b2,b8,b25} introduced energy-aware offloading policies that reduce energy consumption while maintaining low task completion times. Similarly,~\cite{b38} proposed distributed strategies for IoT devices that minimize energy use while preserving execution timeliness. Energy consumption is the primary objective in~\cite{b34}, while~\cite{b75} balances both energy efficiency and delay in UAV-assisted offloading scenarios.

In addition to latency and energy, some models aim to jointly optimize \textit{data rate, bandwidth, and QoS/QoE metrics}. For instance,~\cite{b17} examines trade-offs between latency, bandwidth, and transmission reliability in dynamic edge settings. Study~\cite{b22} focuses on data rate and latency optimization to ensure efficient task handling under heavy network load. To improve user-perceived experience,~\cite{b13} introduces adaptive video quality adjustment that balances bandwidth and latency, while~\cite{b70} incorporates channel access strategies to support delay-sensitive applications.

\begin{table*}[!t]
  \centering

  \caption{Comparison of DRL-Based Offloading Schemes in Hierarchical and Heterogeneous Systems}
  \label{tab:het_comparison}
    \begin{tabular}{|c|c|c|c|p{4cm}|c|}
      \hline
      \makecell[c]{\textbf{Agent}\\\textbf{Type}} &
      \textbf{Ref.} &
      \makecell[c]{\textbf{DRL}\\\textbf{Method}} &
      \makecell[c]{\textbf{Optimization}\\\textbf{Objective}} &
      \makecell[c]{\textbf{Key}\\\textbf{Techniques}} &
      \makecell[c]{\textbf{Computing}\\\textbf{Source}} \\
      \hline\hline

    \multirow{16}{*}{\centering Single Agent} 
        & \cite{b8}  & \multirow{2}{*}{DQN}   & Energy, Latency                    & Distributed RL                      & Local, Edge, Cloud \\ \cline{2-2}\cline{4-6}
        & \cite{b22} &       & Latency, Data Rate                & Centralized and Distributed DRL     & Local, UAV, Cloud \\ \cline{2-6}
        & \cite{b34} & \multirow{6}{*}{DDPG} 
                               & Energy                                & FCNN with Convex Optimization       & Local, Edge, Cloud \\ \cline{2-2}\cline{4-6}
        & \cite{b13} &       & Latency, Bandwidth                & Adaptive Video Quality Transmission & Local, Edge, Cloud \\ \cline{2-2}\cline{4-6}
        & \cite{b14} &       & Latency                              & Social-Relation Clustering          & Local, Edge, Cloud, Vehicles \\ \cline{2-2}\cline{4-6}
        & \cite{b17} &       & Energy, Bandwidth, Latency           & Ornstein-Uhlenbeck Noise            & Local, Edge, Cloud \\ \cline{2-2}\cline{4-6}
        & \cite{b18} &       & Latency                              & Cooperative Edge Processing         & Local, Edge, Cloud \\ \cline{2-2}\cline{4-6}
        & \cite{b21} &       & Latency                              & Prime-Dual DRL Formulation          & Local, Edge, Cloud \\ \cline{2-6}
        & \cite{b51} & \multirow{4}{*}{TD3} 
                               & Latency                              & Edge-Assisted Learning              & Local, Edge, Cloud \\ \cline{2-2}\cline{4-6}
        & \cite{b54} &       & Peak Age of Information              & UAV Scheduling for AoI Minimization & Local, Edge, UAV \\ \cline{2-2}\cline{4-6}
        & \cite{b55} &       & Resource Utilization Efficiency      & Hierarchical SFC Offloading         & Local, Edge, Cloud \\ \cline{2-2}\cline{4-6}
        & \cite{b56} &       & Delay, Energy                     & Mobility-Aware Dynamic Modeling     & Local, Edge \\ \cline{2-6}
        & \cite{b46} & A3C   & Latency                              & Relative Delay-Based Fairness       & Local, Edge, Cloud \\ \cline{2-6}
        & \cite{b44} & \multirow{2}{*}{PPO}  & Latency                              & Distributed PPO with Trajectory Sharing & Local, Edge, Cloud \\ \cline{2-2}\cline{4-6}
        & \cite{b28} &       & Latency                              & Seq2Seq BiGRU-Based Scheduling      & Local, Edge, Cloud \\ \cline{2-6}
        & \cite{b58} & \multirow{2}{*}{SAC}  
                               & Latency, Energy                   & Entropy-Normalized SAC              & Local, UAV \\ \cline{2-2}\cline{4-6}
        & \cite{b57} &       & Service Latency                      & Centralized-Decentralized Agent Mix & Local, Edge, Cloud \\
      \hline

    \multirow{10}{*}{\centering Multi Agent} 
        & \cite{b1}  & \multirow{2}{*}{\makecell[c]{MADQN \\ (DTDE)} } 
                               & Energy, Latency                   & Federated Q-Learning                & Local, Edge, Cloud \\ \cline{2-2}\cline{4-6}
        & \cite{b38} &       & Energy, Task Count                & Independent IIoT Agents             & Local, Edge, Cloud \\ \cline{2-6}
        & \cite{b2}  & \makecell[c]{MADDQN \\ (DTDE)}   
                               & Energy, Latency                   & Trust-Based Secure Offloading       & Local, Edge, Cloud \\ \cline{2-6}
        & \cite{b67} & \makecell[c]{Q-Learning \\ (DTDE)}  
                               & Latency                              & Joint Action Learning   & Local, Edge, AP \\ \cline{2-6}
        & \cite{b25} & \multirow{3}{*}{\makecell[c]{MADDPG \\ (CTDE)}  } 
                               & Latency, Energy                   & LSTM + BRNN State Prediction        & Local, Edge, Cloud \\ \cline{2-2}\cline{4-6}
        & \cite{b30} &       & Latency                              & Centralized Training, Distributed Execution & Local, Edge, Cloud \\ \cline{2-2}\cline{4-6}
        & \cite{b70} &       & Latency, Channel Access           & Prioritized Reward Shaping          & Local, Edge, Cloud \\ \cline{2-6}
        & \cite{b59} & \makecell[c]{MASAC \\ (DTDE)}   
                               & Latency, Success Rate             & Fairness          & Local, Edge \\ \cline{2-6}
        & \cite{b69} & \makecell[c]{MAA3C \\ (DTDE)}   
                               & Delay                                & Knowledge-Driven Learning           & Local, Edge (BS) \\ \cline{2-6}
        & \cite{b75} & \makecell[c]{MAAC \\ (CTDE)}  
                               & Latency, Energy                   & Multi-Objective Formulation         & UAV, Edge \\
      \hline
    \end{tabular}%
  \vspace{-0.75em}
\end{table*}

Table~\ref{tab:het_comparison} summarizes key DRL-based offloading schemes applied in hierarchical and heterogeneous edge computing architectures, detailing agent configurations, optimization objectives, core techniques, and computation sources.

\subsection{Challenges and Limitations}
\subsubsection{Modeling and Representation} 
Several studies have exhibited limitations in state representation, leading to suboptimal decision-making and reduced performance. For instance, \cite{b8} did not incorporate transmission and computation dynamics in their state modeling, resulting in ambiguity in state transitions. Similarly, \cite{b18} omitted temporal information in edge caching decisions, limiting the long-term effectiveness of task processing. In \cite{b21}, the absence of fading channel gain modeling diminished the precision of computational task offloading decisions. Likewise, \cite{b34} neglected to include task dynamics, resource dynamics, and channel variability, key components for adapting to real-world environments, and also failed to consider task arrival rates and server-side waiting times, both of which are crucial for effective scheduling.

Scalability and practical deployment constraints were also overlooked in several studies. For example, \cite{b38} did not account for bandwidth limitations or communication delays between IIoT devices and edge servers, limiting the model's applicability to large-scale deployments. In \cite{b2}, the integration of transmission dynamics in the Double DQN framework lacked clarity, potentially compromising decision accuracy. Similarly, \cite{b25} did not clearly explain how task priority and latency were scaled in the reward function, which could reduce the robustness of the learning process. In \cite{b13}, the adaptation of DDPG for discrete action spaces in mobile VR scenarios was insufficiently explained, and the absence of explicit state transition modeling further impaired decision-making. Moreover, \cite{b14} did not clarify the temporal validity of clustering aggregation or whether the DDPG model updated its state after each aggregation.

Other works also lacked important distinctions. In \cite{b28}, dependent and independent subtasks were not decoupled, potentially limiting the scalability of the task model. \cite{b1} failed to justify the need for uniform state representations across fog zones, despite differing numbers of fog servers and RSUs. For federated learning, this uniformity is critical to support model aggregation, yet the authors did not address this requirement. Additionally, \cite{b75} lacked analytical justification for the observed relationship between increased UAV altitude and task completion rates, even though higher UAV elevation typically incurs greater path loss, which could impair offloading performance.

\subsubsection{Synchronization and Coordination}
Synchronization among agents and networks remains a persistent challenge. For instance, \cite{b8} did not specify a mechanism for synchronizing parallel policy networks, potentially leading to inconsistencies in multi-agent learning. \cite{b38} lacked coordination protocols among IIoT devices, which may limit scalability and increase server-side conflicts. Similarly, \cite{b44} did not address contention management strategies, such as time-division or frequency-division multiplexing, to resolve simultaneous offloading requests, resulting in server bottlenecks. In \cite{b67}, although joint action Q-learning was applied, fairness concerns emerged as agents exhibited selfish behaviors rather than collaborative decision-making. Additionally, \cite{b69} employed A3C with shared weights, but without shared information exchange, which constrained the effectiveness of cooperative multi-agent learning.

\subsubsection{Reward Function and Optimization Trade-offs}
Reward design posed significant challenges in many works. In \cite{b17}, squaring latency and energy terms reduced their contribution when values were less than one, weakening their effect on learning. \cite{b25} encountered difficulty in scaling task priority and latency terms, which may distort optimization. Similarly, the reward function in \cite{b70}, which employed priority weights, did not lead to clear performance benefits, and its impact on latency, energy, or agent-specific results was insufficiently evaluated.

\subsection{Conclusion}
Hierarchical offloading offers a scalable framework for distributing tasks across multi-tier architectures that include cooperative vehicles, edge nodes, fog layers, and cloud infrastructure. By enabling task scheduling across heterogeneous computational resources, it supports low-latency responses for lightweight tasks while accommodating the processing demands of more intensive workloads. Although existing DRL-based strategies have improved system responsiveness and resource efficiency, several unresolved issues remain. Incomplete MDP representations, weak synchronization schemes, reward design flaws, and limited scalability in real-world conditions all present substantial barriers. Addressing these challenges is critical to unlocking the full potential of hierarchical offloading systems in dynamic vehicular and IoT environments.
\section{Lessons Learned, and Open Issues}
\label{sec:lessons}
Based on the findings from the three architectural frameworks, namely centralized, distributed, and hierarchical, we establish a structured set of lessons that synthesizes key examples, underlying intuitions, working hypotheses, and future research directions.
\subsection{Environmental Modeling and Learning Foundation}
\subsubsection{Fully Informed MDP and Multi-Agent POMDP}
Many current studies in VEC adopt MDP or POMDP frameworks but suffer from incomplete state representations. These omissions include key dynamics such as transmission rates, processing delays, resource availability, and caching dependencies, resulting in ambiguous state transitions and suboptimal decision-making. In multi-agent settings, partial observability exacerbates this challenge as agents act on local information, often producing conflicting or redundant actions. Moreover, the curse of dimensionality intensifies with the increasing number of observed wireless channels, making real-time optimization computationally expensive. To address these limitations, future research should prioritize scalable, low-overhead state abstraction techniques, such as dimensionality reduction and task-aware encoding, that support robust and efficient decision-making in both centralized and decentralized environments.
\subsubsection{Handling DAGs, Task, and Data Dependencies}
Computational task offloading in vehicular networks often involves interdependent computations, which can be effectively modeled using DAGs. DAGs help distinguish between dependent subtasks, where outputs from one task are required by another, and independent subtasks that can be processed in parallel. Sequential learning models like Seq2Seq are suitable for handling dependencies, while parallel strategies can optimize independent task execution. Additionally, separating temporal features (e.g., vehicle speed, mobility patterns) from non-temporal ones (e.g., task size, channel state) can enhance learning efficiency. Temporal dynamics are best captured through recurrent models like LSTMs, whereas static features may be processed through standard encoders. Decoupling these feature types allows the learning process to more accurately reflect real-world conditions, thereby improving both adaptability and decision quality in dynamic environments.
\subsubsection{Reward Function Design and Scaling}
Effective reward function design is pivotal in guiding learning behavior, especially when optimizing multiple objectives such as latency and energy. Improper scaling, such as squaring latency or energy, can lead to disproportionate gradients and unbalanced learning, reducing the model's ability to converge toward optimal solutions. Similarly, using raw latency and energy values without appropriate weighting coefficients may cause one objective to dominate the reward, undermining joint optimization efforts. 

Another important consideration is the reliance on relative reward improvements across consecutive time steps. While this can accelerate early learning, it may also lead agents to prioritize short-term gains over long-term system performance, especially in environments with fluctuating traffic or intermittent communication. This short-sightedness increases the risk of agents settling in local optima. To mitigate this, reward design should incorporate more stable indicators, such as moving averages or windowed performance metrics, to distinguish transient variations from systemic trends. Blending short-term responsiveness with long-term goals can produce more balanced policies that generalize better across dynamic conditions. Ultimately, well-calibrated, multi-objective reward functions are key to achieving both performance stability and optimization accuracy in real-world deployments.

Existing DRL-based offloading studies i.e., \cite{b66} ,\cite{b28} ,\cite{b67},\cite{b46} adopt diverse reward formulations, ranging from latency-only objectives to aggregate utility functions that implicitly combine latency, energy, and task completion. However, many of these designs rely on ad hoc scaling or fixed weighting coefficients without systematic justification, which can bias learning toward a dominant objective. In several cases, energy efficiency is evaluated post hoc rather than explicitly encoded in the reward, making it difficult to assess whether reported multi-objective gains reflect true trade-offs or emergent behavior. This observation motivates the need for principled reward scaling and explicit objective balancing, particularly in dynamic vehicular environments.

\subsubsection{Continuous vs. Discrete Action Space Handling}
Deterministic policy gradient methods such as DDPG, PPO, and TRPO are well-suited for continuous action spaces, as they output a single deterministic action rather than a probability distribution. However, applying these methods to discrete action spaces i.e., \cite{b15}, \cite{b29}, and \cite{b31}  introduces challenges, primarily because they are inherently designed for continuous outputs and rely on added noise (e.g., Ornstein-Uhlenbeck or Gaussian noise) for exploration . In discrete settings, this noise can cause frequent switching between actions, leading to unstable learning dynamics and inefficient convergence. A key concern is achieving balanced exploration across all available actions. Without careful control, noise injection can result in biased action selection, limiting the agent's ability to discover optimal policies. Moreover, the misalignment between continuous output generation and discrete action requirements may degrade performance unless specifically addressed. Designing effective mappings or hybrid strategies to bridge this gap is a promising research direction. For instance, discretization layers or quantization techniques could allow deterministic methods like DDPG to operate  in discrete domains. Exploring such solutions is essential for adapting continuous-control algorithms to real-world scenarios where discrete actions, such as task selection, resource allocation, or server assignment, are common.

\subsubsection{Network Architecture Selection Guidance}
In essence, VEC architecture selection reflects a trade-off between network dynamics and problem coupling, balancing scalability, latency, and coordination complexity. A centralized MEC architecture \cite{b42,b45} is effective in small-scale VEC scenarios with bounded mobility, such as smart campuses or industrial parks, where a single MEC server co-located with the base station serves all vehicles. Short communication distances enable low latency and reduced vehicle energy consumption, while global system observability supports centralized optimization. However, scalability limitations, communication channel congestion, computational overhead and large number of requests being in waiting queue hinder its applicability in large-scale or highly dynamic vehicular networks. As the number of users and the spatial extent of edge computing demand increase, the architecture selection must shift toward distributed computing systems \cite{b6,b12,b15,b54}; consequently, the need for multi-edge-supported architectures emerges. For instance, in typical urban environments with dense road networks and multiple intersections-either linear corridors or circular junctions, the high mobility and concentration of vehicular nodes drive the deployment of roadside edge servers at strategic locations. These distributed MEC servers collectively provide substantial computational resources, reduce uplink congestion, and ensure low-latency task execution for a large population of vehicles, which would be infeasible under a single centralized MEC architecture. Moreover, in heterogeneous scenarios such as those considered in \cite{b58a}, where urban, hotspot, and highway environments coexist, the diversity in mobility patterns, traffic density, and service requirements highlights the necessity of a multi-agent distributed computing system. Such architectures naturally support collaborative learning and decentralized decision-making, enabling effective coordination among agents while maintaining scalability and responsiveness across heterogeneous VEC environments.

However, in many cases, vehicular nodes require high computational support that cannot be efficiently handled by distributed edge servers alone. In such scenarios, the architecture must incorporate a hierarchical and heterogeneous design \cite{b1,b2,b57}, where distributed MEC nodes are augmented by higher-tier, high-capacity computing resources, such as cloud servers, satellite-assisted MEC, or other centralized high-performance MEC platforms. A representative example is large-scale urban autonomous driving, where roadside MEC servers handle latency-sensitive tasks such as object detection, cooperative perception, and local trajectory planning, while computation-intensive workloads such as high-definition map construction, long-horizon route optimization, and large-scale model training are offloaded to upper-tier cloud or satellite-assisted MEC layers. This hierarchical coordination enables low-latency local responsiveness while preserving the ability to process computationally demanding tasks that exceed the capacity of distributed edge nodes.

\subsection{Multi-Agent Coordination, Synchronization, and Heterogeneity}
\subsubsection{Multi-Agent Coordination}
One of the key lessons from recent studies i.e., \cite{b12},\cite{b43}, \cite{b20},\cite{b64},  is the critical importance of synchronization and coordination in multi-agent systems and distributed networks. Inadequate synchronization often leads to inefficiencies and scalability challenges, especially as the number of agents or system complexity increases. Misaligned agent actions can cause resource contention and bottlenecks in environments where multiple users attempt to offload tasks or access shared communication channels, ultimately degrading overall system performance. For example, in VEC or UAV-assisted MEC systems, static resource allocation can result in significant inefficiencies. Consider a typical FDMA scenario with three users, each allocated a separate frequency band: if one user completes its task early, the unused bandwidth remains idle unless dynamically reassigned. This rigidity causes resource underutilization and increased latency for others with higher task loads. As the number of agents grows, such inefficiencies compound, intensifying queuing delays, degrading fairness, and raising scalability concerns. To address these challenges, systems must adopt contention-aware scheduling, dynamic resource allocation policies, and adaptive coordination mechanisms. 

Furthermore, the integration of DRL with federated learning has emerged as a promising paradigm for heterogeneous VEC systems, as it enables privacy-preserving and scalable learning under non-IID (Non-Independent and Identically Distributed) data distributions and dynamic vehicular mobility. Representative studies, such as \cite{b64a},\cite{b64b}, demonstrate that federated parameter aggregation can improve cache hit rate, delay, and training stability without sharing raw data, while federated multi-agent learning i.e., \cite{b64c} shows effectiveness for global AoI minimization in MEC environments. However, a key bottleneck remains that collaboration among agents is largely implicit, achieved through parameter averaging or shared global rewards, rather than explicit coordination of coupled actions or resource contention. Consequently, federated DRL alone does not fully resolve decision coupling, non-stationarity, or competitive interactions among agents sharing wireless, caching, or computation resources, which may limit its effectiveness in strongly coupled and highly heterogeneous VEC scenarios. Future research should therefore explore hybrid federated–coordinated DRL frameworks that preserve decentralized training and privacy while incorporating lightweight coordination mechanisms (e.g., centralized or hierarchical critics, coordination-aware rewards, or selective information sharing) to explicitly handle inter-agent coupling in VEC systems.

\subsubsection{Synchronous Learning in Multi-Objective Multi-Agent Systems}
In multi-objective multi-agent systems, decision dependencies among agents necessitate synchronized learning and coordination. Without proper synchronization, agents may act independently and asynchronously, resulting in inefficiencies such as increased latency, idle tasks, or misallocated resources. For instance, if a base station offloads a task to a UAV whose learning process is unsynchronized or delayed, the task may remain unprocessed, causing performance degradation across the system. 

To mitigate these issues, one approach is to implement a global time step that aligns decision-making across agents while incorporating hierarchical structures and longer decision horizons. Alternatively, asynchronous learning mechanisms can be enhanced with adaptive delay strategies, where agents update their policies based on local environmental conditions, such as queue length, energy level, or task arrival rate. For example, a UAV with high task arrival frequency and mobility may update its policy more frequently, while another UAV operating in a low-activity zone may defer updates until specific thresholds are met. These techniques strike a balance between synchronization and responsiveness, improving coordination, scalability, and overall system performance in dynamic environments.

Moreover, multi-objective optimization fundamentally depends on the reward function design. As discussed in our survey, several DRL-based offloading studies infer multi-objective performance primarily from evaluation metrics reported after training, rather than explicitly incorporating trade-offs within the reward formulation. This practice makes it difficult to determine whether reported gains in energy and latency reflect a genuine balance or the dominance of a single objective. In particular, some centralized offloading studies optimize latency-focused objectives while leaving energy efficiency only implicitly addressed through the reward design or secondary constraints, thereby risking the under evaluation of energy–latency efficiency (e.g., \cite{b3}). Similarly, other works rely on aggregate utility or cost metrics dominated by latency terms, without disentangling the relative contributions of energy, which further obscures multi-objective trade-offs (e.g., \cite{b7}). In contrast, reward structures that jointly integrate latency and energy such as weighted formulations or latency-constrained rewards enhanced with explicit energy penalties, can ensure that neither objective is unintentionally overlooked. However, our analysis indicates that such explicit balancing mechanisms are not consistently adopted across existing studies. In particular, methods that strongly prioritize latency through hard constraints can improve task completion, but they often do so at the cost of energy efficiency. When latency dominates the reward signal, energy considerations tend to receive less attention, which has been observed in several distributed and centralized DRL frameworks reviewed in this survey (e.g., \cite{b5, b11}).

\subsubsection{Challenges in Heterogeneous Multi-Agent Environments}
In heterogeneous multi-agent systems, agents often operate with varying state and action spaces due to differences in their local observations, functional roles, or hardware capabilities. For instance, a UAV-based server encounters dynamic wireless conditions and mobility constraints; a terrestrial MEC node manages structured urban traffic with low-latency demands; and a satellite server deals with sparse, long-range sensing and delayed feedback. These disparities result in diverse state encodings and decision types across agents. While unified representations could simplify coordination, enforcing strict standardization is often infeasible in decentralized systems. Therefore, future work must focus on adaptive mechanisms that accommodate heterogeneity, such as modular state encoders, flexible action abstractions, and semantic alignment techniques. These strategies can enable policy compatibility and support effective collaborative learning, especially in systems where agents pursue multi-objective or role-specific goals.
\subsection{Algorithmic Integration , and Evaluation}

\subsubsection{Game-Theoretic Integration Challenges}
Most existing Game Theory-DRL frameworks employ hierarchical or sequential architectures, where game theory sets up the structure and DRL learns within fixed payoff models. This design restricts adaptability in dynamic and asynchronous systems. A promising alternative is a co-adaptive hierarchical game architecture in which high-level decision making governs the evolution of the payoff structure, while low-level DRL policies perform fine-grained exploration and control. For example, in the context of edge computing, the upper layer can adapt payoff mechanisms, cluster sizes, and cluster composition based on global network conditions, whereas the lower layer learns task or user scheduling decisions, and resource allocation under these evolving rules. The interaction between these layers allows the underlying game equilibrium to evolve in response to dynamics such as congestion and resource availability. This hierarchical co-adaptation enables the system to track time-varying equilibria, thereby improving both real-time responsiveness and long-term system stability..


\subsubsection{Reliable Baseline Comparisons and Evaluation}
Robust baseline comparisons are critical for fair and meaningful evaluation in RL research. Benchmarking new algorithms against well-established baselines provides a clearer understanding of their effectiveness across diverse scenarios and ensures their practical applicability. However, many studies suffer from incomplete or inconsistent comparisons, which can distort results and obscure the real contributions of new methods. In MARL, the lack of dedicated multi-agent baselines further complicates evaluations. Without accounting for the unique dynamics and coordination challenges in MARL settings, comparisons against single-agent benchmarks may yield biased assessments. Additionally, the omission of essential performance metrics, such as energy consumption, task efficiency, and system throughput, limits the ability to evaluate solutions holistically. Incomplete reporting of experimental conditions and results reduces generalizability and hinders reproducibility.

Furthermore, unstable learning, manifested as reward fluctuations or inconsistent convergence, can weaken the credibility of reported outcomes. Fixed exploration rates or premature convergence patterns may fail to adapt to dynamic environments, reducing the system's robustness. These issues underscore the need for comprehensive baselines, dynamic exploration strategies, and multi-metric evaluation frameworks to enable more accurate, scalable, and reliable assessments of DRL algorithms in complex settings.
\subsection{Deployment Practices}

\subsubsection{Real-World Deployment}
Most DRL-based offloading solutions are evaluated in small-scale, synthetic simulation environments, often overlooking the computational and communication overheads associated with real-time inference and the constraints of in-vehicle or on-device deployment. In addition, many existing approaches do not adequately address high-dimensional action spaces, such as frequency-selective channel allocation, where the number of discrete actions increases rapidly with the number of users and available sub-channels. This action-space explosion significantly complicates policy optimization, particularly in user-centric and highly dynamic vehicular environments. Collaborative decision-making among agents further intensifies these challenges, as coordination must simultaneously consider bandwidth limitations, latency sensitivity, and restricted computational resources. Consequently, future research should explore scalable solutions including lightweight policy architectures, model compression, distributed or federated DRL frameworks, and hardware-in-the-loop testbeds to rigorously assess real-world feasibility. A particularly promising direction is the development of generalized DRL models that can adapt to varying numbers of agents, enabling vehicles to dynamically join or leave the system without retraining from scratch.

An effective strategy to address scalability in DRL-based vehicular edge computing is zone-wise agent deployment, where the road network is partitioned into spatially or traffic-aware zones (e.g., intersections, road segments, or congestion hotspots). Within each zone, agents operate under a shared partially observable Markov decision process (POMDP) based on local observations such as channel conditions, queue states, and traffic density, thereby reducing state and action dimensionality and limiting coordination overhead. By sharing a common POMDP formulation, policies can be parameter-shared and reused across zones, improving sample efficiency and generalization under fluctuating agent populations. Scalability can be further enhanced through zero-shot and few-shot generalization techniques, including transfer learning, meta-learning, and semantic representation learning, which allow agents to infer suitable offloading and resource allocation decisions when entering previously unseen environments. When combined with CTDE, zone-level policies can leverage global information during training while remaining lightweight enough for real-time, on-device inference, making this hierarchical and modular design well suited for large-scale, user-centric vehicular edge computing systems.
\subsubsection{Scalability and Model Generalization}
Scalability challenges in large-scale DRL-based VEC networks can be addressed through zero-shot deployment, supported by training strategies such as transfer learning, meta-learning, and environmental semantic representation learning. Transfer learning reuses knowledge learned from a specific source environment to deploy or fine-tune a model in a new but related environment, thereby reducing training overhead. Meta-learning, in contrast, pretrains the model across multiple diverse environments so that shared experience enables rapid adaptation when a previously unseen environment is encountered. In vehicular settings, semantic abstractions such as line-of-sight (LOS) and non-line-of-sight (NLOS) channel characteristics, interference regimes, and resource availability patterns can be incorporated during pretraining to capture environment-invariant features. For example, a vehicular agent entering a new geographical zone where it was not explicitly trained may leverage these learned semantic representations to infer suitable offloading and resource allocation decisions upon deployment, enabling zero-shot or few-shot operation without retraining the entire system. Collectively, transfer learning could provide reusable prior knowledge, meta-learning may enable rapid adaptation, and semantic representation learning can promote environment invariance, which together would support scalable generalization and facilitate zero-shot deployment in large-scale VEC systems.

\subsubsection{Real-time Inference Challenges}
Issues such as model convergence in highly dynamic vehicular environments can be partially alleviated through collaborative learning, multi-agent coordination, and improved generalization techniques. However, a more critical challenge arises at the inference stage, where individual vehicles are often constrained by limited computational and memory resources, making it difficult to execute complex DRL models in real time. In addition, when decision-making depends on edge or network-level information, inference latency is further increased due to the need to transmit state information to edge servers and wait for response decisions, which may violate the strict latency requirements of vehicular applications.

One promising direction to address these limitations is to decompose the overall decision problem into hierarchical subproblems, for example through a Stackelberg game formulation. In such a framework, a network controller or edge server can act as the leader, handling high-dimensional and global decision components, while lightweight user agents operate as followers and make local decisions using simplified policies. Moreover, a distributed multi-agent learning architecture can be employed, where the network controller and edge nodes collaborate during training and subsequently disseminate specialized policies tailored to different user contexts. After convergence, the leader may only broadcast compact subspace decisions or guidance signals, enabling vehicles to perform low-latency, real-time inference locally using lightweight models, while reducing both the amount of state information transmitted during request phases and the volume of decision information exchanged in response interactions with the edge server.

\section{Conclusion and Future Research Roadmap}
\label{sec:conclusion}
\subsection{Core Challenge for Centralized Offloading Architectures: Scalability and Robustness}
Centralized DRL-based offloading architectures offer a structured and globally informed decision-making framework, which enables effective optimization of latency, energy consumption, and resource utilization under a unified MDP formulation. The availability of global state information simplifies coordination and policy learning; therefore, these architectures are attractive for moderate-scale vehicular edge networks. However, the fundamental challenge of centralized architectures lies in scalability and robustness. For instance, along with vehicular density, task heterogeneity, and mobility, an increase in the number of decision variables i.e., multiple users offloading at a time leads to an exponential growth in the decision space. Consequently, centralized controllers or agent face computational bottlenecks in deriving optimal decisions, incur excessive communication and computational overhead, and remain vulnerable to single-point failures. Moreover, simplified state and reward representations commonly adopted in centralized models limit their ability to generalize to highly dynamic real-world environments, underscoring the need for more realistic modeling and scalable learning mechanisms.

\subsection{ Core Challenge for Distributed Offloading Architectures: Coordination under Non-Stationarity} Distributed offloading architectures significantly enhance scalability, fault tolerance, and responsiveness by enabling multiple edge servers or agents to make localized decisions based on partial observations. DRL and MARL frameworks applied in this paradigm demonstrate strong potential in handling mobility, task migration, and dynamic resource allocation across spatially distributed edge nodes. Nevertheless, the core challenge in distributed architectures is effective coordination under non-stationary environments. Independent learning agents continuously alter the environment dynamics for one another, leading to unstable training, suboptimal convergence, and resource contention. Inadequate synchronization, poorly calibrated reward structures, and inconsistent benchmarking further exacerbate these challenges; therefore, robust coordination mechanisms and principled evaluation frameworks remain critical open problems.

\subsection{ Core Challenge for Hierarchical Offloading Architectures: Cross-Layer Consistency and Complexity}

Hierarchical offloading architectures aim to bridge the gap between centralized and distributed architecture by decomposing decision-making across multiple tiers, such as vehicles, edge servers, fog nodes, UAVs, and cloud platforms. This paradigm enables flexible task placement, improved load balancing, and enhanced adaptability to heterogeneous resources and application requirements. However, its primary challenge lies in maintaining cross-layer consistency amid increased system complexity. Misalignment between high-level coordination policies and low-level execution decisions can result in inefficiencies, delayed adaptation, or unstable learning. Additionally, designing coherent MDP formulations, coordinated reward strategies, and synchronization mechanisms across tiers introduces substantial modeling and implementation complexity and consequently highlights the need for more robust hierarchical learning and scalable coordination strategies.

\subsection{Research Roadmap Toward Scalable and Robust VEC Intelligence}

\begin{figure*}[!t]
  \centering
  \includegraphics[width=0.9\textwidth]{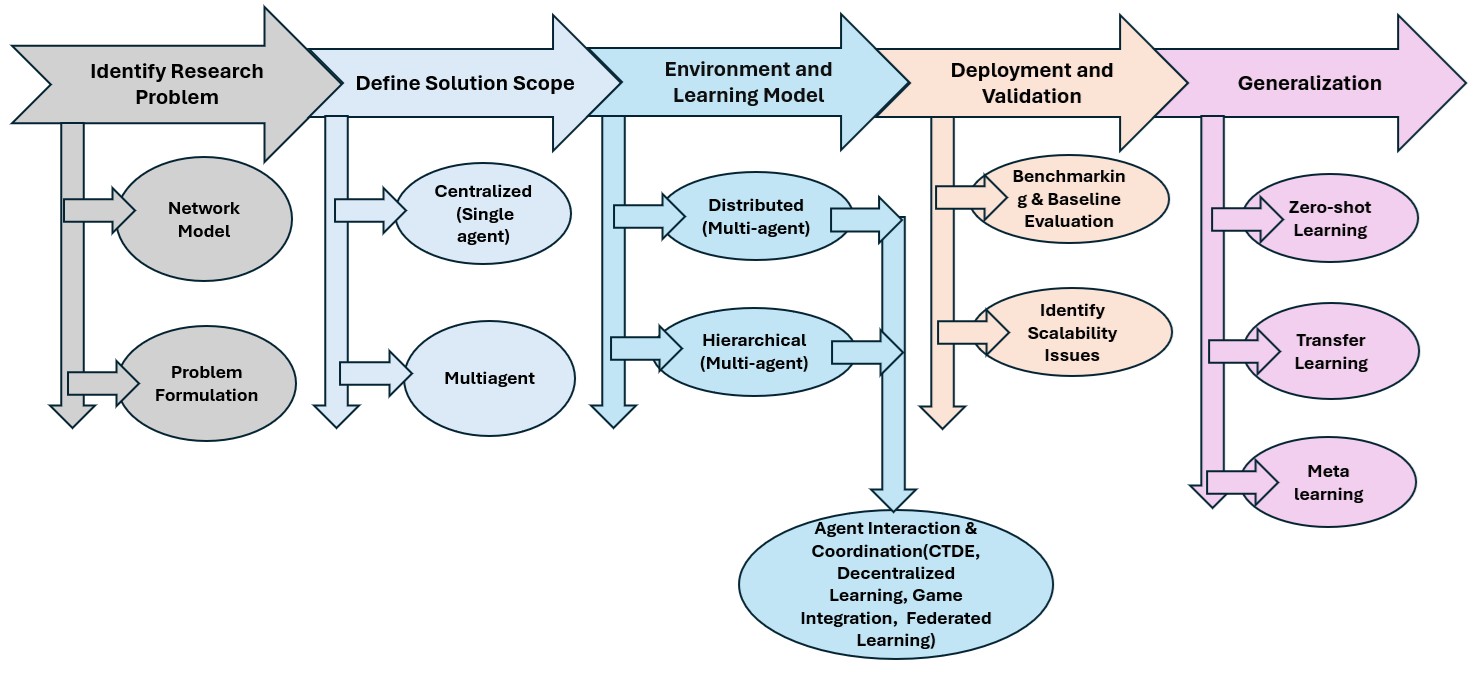}
    \caption{Roadmap of Research.}
    \label{fig:Roadmap}
\end{figure*}

To consolidate the insights drawn from centralized, distributed, and hierarchical offloading paradigms, Fig. \ref{fig:Roadmap} presents a high-level research roadmap. 
The roadmap progresses from identifying the research problem and system model, through defining the solution scope using either centralized or multi-agent learning paradigms, to environment modeling and agent interaction design. It further emphasizes deployment-oriented evaluation through benchmarking and scalability analysis, and highlights generalization mechanisms such as transfer learning, meta-learning, and zero-shot learning as key enablers for robust real-world adoption.

\subsection{Conclusion}
This survey has highlighted the growing role of DRL in addressing the challenges of VEC, where intelligent computational task offloading, adaptive decision-making, and resource coordination are crucial in dynamic, distributed environments. Our analysis reveals that scalable and efficient DRL solutions must incorporate several core design principles: informative and uniform state representations, especially in multi-agent and heterogeneous systems; robust reward function design that balances short-term performance with long-term system goals; and comprehensive baseline comparisons that reflect both centralized and decentralized learning scenarios. In multi-agent settings, effective coordination hinges on synchronization mechanisms, shared learning signals, and adaptive training strategies that consider agent heterogeneity and dynamic task dependencies. For scalable deployment in real-world vehicular systems, future DRL architectures must also account for communication delays, computational constraints, and high-dimensional action spaces, while remaining lightweight and generalizable to varying network topologies and traffic conditions. Ultimately, addressing these challenges will enable the development of more resilient, cooperative, and adaptive DRL frameworks for VEC, paving the way toward intelligent, real-time decision-making in next-generation transportation and edge computing infrastructures.

\end{document}